\documentclass[a4paper,fleqn]{cas-dc}

\usepackage[numbers]{natbib}
\usepackage{graphicx}
\usepackage{amsmath}
\usepackage{amsthm}
\usepackage{booktabs}
\usepackage{mathrsfs}
\usepackage{multirow}
\usepackage{colortbl}
\usepackage{booktabs}

\usepackage{bm}
\usepackage{color}
\usepackage{subfigure}
\usepackage{mathrsfs}
\usepackage{multirow}
\usepackage{threeparttable}

\usepackage{arydshln}
\usepackage[ruled,vlined,linesnumbered]{algorithm2e}

\newcommand{\bw}        {{\bf w}}

\def\tsc#1{\csdef{#1}{\textsc{\lowercase{#1}}\xspace}}
\tsc{WGM}
\tsc{QE}

\begin{document}
\let\WriteBookmarks\relax
\def\floatpagepagefraction{1}
\def\textpagefraction{.001}

\shorttitle{Privacy Preserved Blood Glucose Level Cross-Prediction: An Asynchronous Decentralized Federated Learning Approach}    

\shortauthors{Chengzhe Piao {\em et al. }}  

\title [mode = title]{Privacy Preserved Blood Glucose Level Cross-Prediction: An Asynchronous Decentralized Federated Learning Approach}  



%

\author[1]{Chengzhe Piao}[orcid=0000-0003-0494-5098]
\ead{chengzhe.piao.21@ucl.ac.uk}
\credit{Conceptualization, Methodology, Software, Investigation, Writing – original draft}
\author[2]{Taiyu Zhu}
\ead{taiyu.zhu@psych.ox.ac.uk}
\credit{Data curation,  Validation, Formal analysis, Writing – review \& editing}
\author[1]{Yu Wang}
\ead{yu.w.23@ucl.ac.uk}
\credit{Data curation,  Validation, Formal analysis, Writing – review \& editing}
\author[3,4]{Stephanie E Baldeweg}
\ead{s.baldeweg@ucl.ac.uk}
\credit{Conceptualization, Writing – review \& editing}
\author[1]{Paul Taylor}
\ead{p.taylor@ucl.ac.uk}
\credit{Resources, Writing – review \& editing}
\author[5]{Pantelis Georgiou}
\ead{pantelis@imperial.ac.uk}
\credit{Data curation, Resources, Writing – review \& editing}
\author[6]{Jiahao Sun}
\ead{sun@flock.io}
\credit{Conceptualization, Writing – review \& editing}
\author[7]{Jun Wang}
\ead{junwang@cs.ucl.ac.uk}
\credit{Conceptualization, Writing – review \& editing, Supervision}
\author[1]{Kezhi Li}
\cormark[1]
\ead{ken.li@ucl.ac.uk}
\credit{Conceptualization, Writing – review \& editing, Project administration, Supervision, Resources, Funding acquisition}   
         
\affiliation[1]{organization={Institute of Health Informatics, University College London},
                city={London},
                postcode={NW1 2DA}, 
                country={UK}}
\affiliation[2]{organization={Department of Psychiatry, University of Oxford},
                city={Oxford},
                postcode={OX3 7JX}, 
                country={UK}}
\affiliation[3]{organization={Department of Diabetes \& Endocrinology, University College London Hospitals},
            city={London},
            postcode={NW1 2PG}, 
            country={UK}}
\affiliation[4]{organization={Centre for Obesity \& Metabolism, Dept of Experimental \& Translational Medicine, University College London},
            city={London},
            postcode={WC1E 6JF}, 
            country={UK}}
\affiliation[5]{organization={Centre for Bio-Inspired Technology, Department of Electrical and Electronic Engineering, Imperial College London},
            city={London},
            postcode={SW7 2AZ}, 
            country={UK}}

\affiliation[6]{organization={FLock.io},
            city={London},
            postcode={WC2H 9JQ}, 
            country={UK}}
\affiliation[7]{organization={Department of Computer Science, University College London},
            city={London},
            postcode={WC1E 6EA}, 
            country={UK}}












\cortext[1]{Corresponding author}



\begin{abstract}
Newly diagnosed Type 1 Diabetes (T1D) patients often struggle to obtain effective Blood Glucose (BG) prediction models due to the lack of sufficient BG data from Continuous Glucose Monitoring (CGM), presenting a significant ``cold start'' problem in patient care. 
Utilizing population models to address this challenge is a potential solution, but collecting patient data for training population models in a privacy-conscious manner is challenging, especially given that such data is often stored on personal devices. Considering the privacy protection and addressing the ``cold start'' problem in diabetes care, we propose ``GluADFL'', blood Glucose prediction by Asynchronous Decentralized Federated Learning. 
We compared GluADFL with eight baseline methods using four distinct T1D datasets, comprising 298 participants, which demonstrated its superior performance in accurately predicting BG levels for cross-patient analysis. 
%
Furthermore, patients’ data might be stored and shared across various communication networks in GluADFL, ranging from highly interconnected (e.g., random, performs the best among others) to more structured topologies (e.g., cluster and ring), suitable for various social networks. 
%
%
The asynchronous training framework supports flexible participation. By adjusting the ratios of inactive participants, we found it remains stable if less than 70\% are inactive.
Our results confirm that GluADFL offers a practical, privacy-preserving solution for BG prediction in T1D, significantly enhancing the quality of diabetes management.

\end{abstract}


%

\begin{keywords}
Federated Learning \sep
Blood Glucose Prediction \sep
Type 1 Diabetes \sep
Cross-Patient Analysis
\end{keywords}

\maketitle










\section{Introduction}
\label{sec:introduction}
Blood Glucose (BG) prediction \cite{DBLP:conf/ijcai/ContrerasBBVO18, DBLP:journals/sensors/MontaserDB21, DBLP:journals/iotj/ZhuKDHLG23} is indispensable for individuals with Type 1 Diabetes (T1D), as it enables proactive management of BG levels \cite{10313965, 10330994}, thus preventing potential hyperglycemia and hypoglycemia and mitigating complications. A common challenge faced by many, especially those newly diagnosed or recently initiated on Continuous Glucose Monitoring (CGM) devices, is the lack of sufficient BG trajectory data, leading to a ``cold start'' issue in developing accurate prediction models. This problem is particularly acute for critically ill patients in intensive care, where timely and effective BG management is crucial \cite{desgrouas2023insulin}.
Given that population models can address the cold start problem \cite{DBLP:conf/embc/MohebbiJHCTJBM20}, the emergence of Federated Learning (FL) might offer a promising solution to this challenge, especially in gathering population features while regarding data privacy \cite{DBLP:journals/titb/LakhanMNMTVAW23, DBLP:conf/aistats/McMahanMRHA17}.

\begin{figure*}[tb]
\centering
\includegraphics[width=0.8\linewidth]{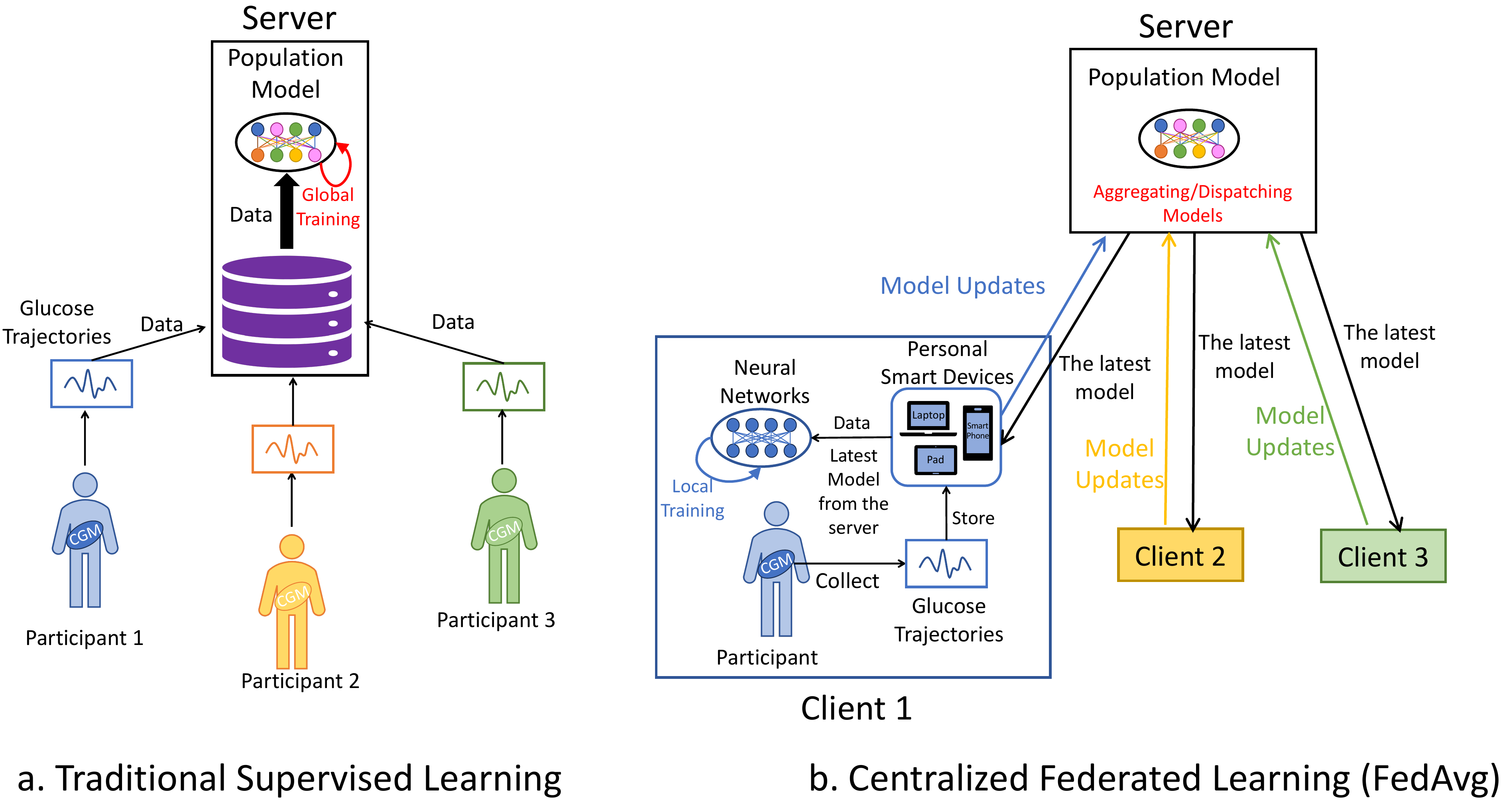}
\caption{ 
Population models for glucose prediction: a) Traditional Supervised Learning, which centralizes patients' data on a server to create models by combining data, and b) Centralized Federated Learning (FedAvg, \cite{DBLP:conf/aistats/McMahanMRHA17}), which utilizes data on personal devices. With FedAvg, only model updates, not raw data, are sent to the server, which then aggregates these updates to improve the population model. This approach maintains data privacy by keeping patient data localized while still benefiting from collective insights. Nevertheless, it has scalability and efficiency challenges in managing large-scale data.}
\label{fig:present}
\end{figure*}

Specifically, employing the integration of data from multiple individuals into population models, FL enables the utilization of patient data on mobile devices or databases in hospitals in a privacy-preserving manner. By directly leveraging the data in the place where it is generated, FL allows for the development of robust prediction models without compromising patient privacy. 
However, applying FL in the cold start problem triggers some concerns:
\begin{itemize}
    \item \textbf{Target Differences}: 
    In a setting similar to zero-shot learning \cite{DBLP:journals/pami/XianLSA19}, patients can be categorized as \textit{seen} or \textit{unseen} based on whether their data has been used to train the model. 
    Patients who have been previously observed (\textit{seen} patients) contribute their data to FL training with the expectation of obtaining personalized models. These personalized models are designed to retain the unique behavioral patterns of individual patients. Conversely, patients who have not been previously observed (\textit{unseen} patients) seek a population model by leveraging seen patients' data. The population model provides them with a foundational starting point, and it is defined as models that encapsulate the general patterns observed across a broad patient cohort (refer to Figure \ref{fig:present}).

    \item \textbf{Overhead and Latency}:
Learning population models in FL necessitates collecting extensive data from patients. However, scalability and efficiency challenges in managing large-scale data, especially in centralized FL (see Figure \ref{fig:present}b), are notable due to increasing overhead and latency with more participant devices. These issues are crucial to resolve for practical healthcare applications.
    \item \textbf{Device Limitations}: Patients with basic smartphones might face longer training times due to limited computational resources. Additionally, frequent phone usage can hinder continuous participation in synchronous FL training, reducing overall efficiency. 
    \item \textbf{Social Preferences}: Some patients may willingly engage in FL training and allow their devices to communicate broadly, while others may prefer limiting interactions to their social circles.
\end{itemize}


\begin{figure*}[tb]
\centering
\includegraphics[width=0.8\linewidth]{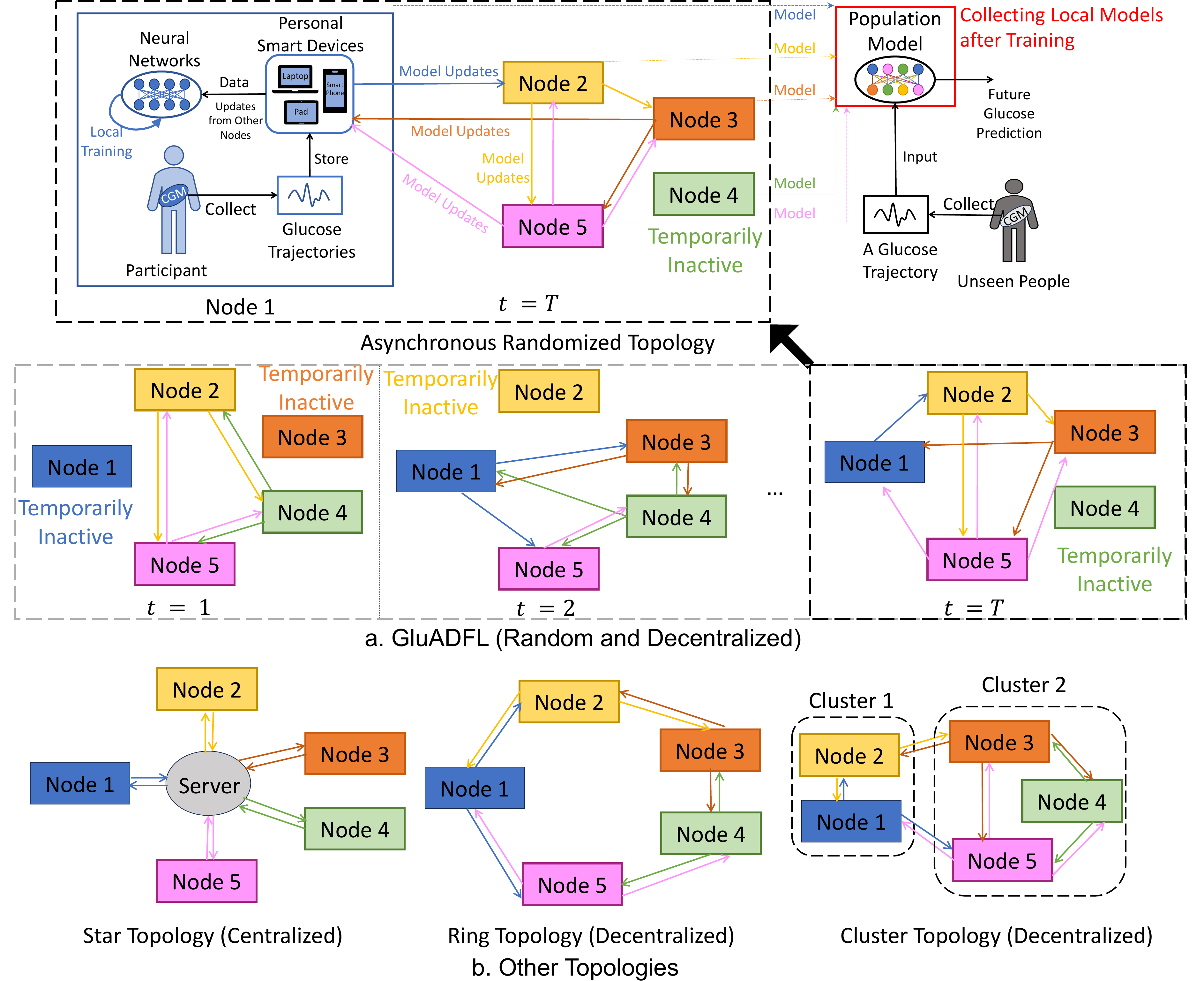}
\caption{A Glucose management system using Asynchronous Decentralized Federated Learning (GluADFL), tailored for Type 1 Diabetes (T1D) management. Figure a showcases the GluADFL employing randomized communication graphs, highlighting the dynamic nature of node interactions for distributed learning. Nodes can mirror the participants (seen patients) attending the training. In these nodes, through CGM devices, glucose data is collected and processed on local devices for model training. Engaging in the GluADFL process, these devices collaboratively develop a population model to enhance BG management for both seen patients with ample CGM data and unseen patients lacking sufficient data. Figure b presents other three distinct communication graph topologies: star, ring, and cluster. Star topology is only for centralized FL (FedAvg \cite{DBLP:conf/aistats/McMahanMRHA17}). Random, ring and cluster can be utilized in GluADFL, each catering to different operational efficiencies and privacy considerations within the network. These topologies reflect diverse participant preferences for data sharing and communication. }
\label{fig:model}
\end{figure*}

In order to address the cold start towards unseen patients, 
the primary objective of FL is to develop a population model that leverages data from seen patients while safeguarding their privacy (Figure \ref{fig:present}). This population model is intended to be applicable to both \textit{seen} and \textit{unseen} patients, which sets it apart from most existing work that focuses on personalized BG prediction. However, in this study, we also introduce a ``personalized from population'' approach that further fine-tunes the population model to better suit each patient's unique characteristics. Specifically,
seen patients can leverage their mobile devices such as smart phones to engage in FL training. Such devices are equipped with storage for CGM sensor data, and tools like TensorFlow Lite support on-device training\footnote{https://www.tensorflow.org/lite/examples/on\_device\_training/overview}. After developing the population model, seen patients can further optionally personalize the model with their data, i.e., ``personalized from population''. Meanwhile, unseen patients can initially use the population model with a ``warm'' start. They can even subsequently fine-tune it as they collect more personal data. Throughout this process, the privacy of personal data is safeguarded, as it remains confined to the individual's device.

Then, given other concerns, a centralized FL structure fails to offer viable solutions.
Therefore, we propose ``GluADFL'', a Glucose management system using Asynchronous Decentralized FL.
Removing the central server allows our method to be scalable for large-scale data applications.
This system accommodates various communication topologies (ring, cluster, random), catering to diverse social needs. By enabling asynchronous training, it mitigates the impact of ``slow'' and ``busy'' devices, allowing broader and more convenient participation in the training process.
Meanwhile, as demonstrated in the BG level prediction challenge 2020 \cite{DBLP:conf/ecai/MarlingB20}, Long Short-Term Memory (LSTM) models \cite{DBLP:conf/ecai/BevanC20, yang2020multi} have proven their efficacy. Besides, well-developed APIs for LSTM models, such as TensorFlow or PyTorch, are widely supported across a broad range of mobile devices. Conversely, other models, such as Neural Basis Expansion Analysis for interpretable Time Series (N-BEATS, \cite{DBLP:conf/iclr/OreshkinCCB20}), Neural Hierarchical interpolation for Time Series (NHiTS, \cite{DBLP:conf/aaai/ChalluOORCD23}) or customized transformers \cite{DBLP:conf/nips/VaswaniSPUJGKP17}, have yet to receive similar support.
Therefore, we choose LSTM as the prediction method for GluADFL due to its reliable performance and well developed APIs on mobile devices.

To evaluate GluADFL with LSTM, our experiments consider four diverse datasets, involving 298 participants. Our findings are summarized as follows: 
\begin{itemize}
    \item The LSTM-based population model, trained through GluADFL, possesses the capability to cross-predict, meaning it can accurately predict BG levels for unseen patients. Additionally, fine-tuning this model with data from seen patients results in personalized models that significantly outperform those created from random initialization. Along with the safe privacy, this improvement in BG prediction is an attractive incentive for seen patients to participate in FL training to refine their personalized models.
    \item We evaluate the performance of GluADFL across different communication topologies, i.e., ring, cluster, and random, to cater to various social needs. The random topology emerges as relatively the most effective, showing a tendency to converge with the best performance. To assess the robustness of GluADFL in asynchronous environments, we experiment with varying inactive node ratios. This approach simulates the impact of different levels of participant inactivity on the training efficiency of the population model. We observe that in a random-based GluADFL setting, the training framework's effectiveness is significantly impacted only when the proportion of inactive nodes exceeds $70\%$.

    \item We compare the cross-patient BG prediction capabilities of GluADFL with eight distinct methods. When training population models using seen patients' data via traditional supervised learning, the LSTM-based approach outperforms other methods such as linear regression, eXtreme Gradient Boosting (XGBoost \cite{DBLP:conf/kdd/ChenG16}), N-BEATS, \cite{DBLP:conf/iclr/OreshkinCCB20}, and NHiTS  \cite{DBLP:conf/aaai/ChalluOORCD23}, validating LSTM as a practical choice for population modeling.
    Furthermore, the GluADFL-based models achieve comparable performance to supervised learning and centralized FL \cite{DBLP:conf/aistats/McMahanMRHA17}, and they are superior to the meta-learning framework \cite{DBLP:conf/icml/FinnAL17, DBLP:journals/corr/LiZCL17} without fine-tuning. This highlights the robust cross-predictive capability of GluADFL-based LSTM population models.
\end{itemize}

\section{Related Work}
FL has significantly evolved since its initial concept with FedAvg \cite{DBLP:conf/aistats/McMahanMRHA17}, a centralized model involving a server coordinating multiple clients. FedAvg's core principle is to train models directly on devices where data originates, avoiding the need for centralized data gathering and enhancing privacy protection.

Decentralized FL has further evolved from this concept by eliminating the central server, enabling broader participation. Various researchers have proposed innovative decentralized FL frameworks. For instance, He \emph{et al.} \cite{DBLP:conf/aaai/0001CBAA22} introduced SpreadGNN, a multi-task federated framework adept at handling partial labels without a central server. Shi \emph{et al.} \cite{DBLP:conf/icml/Shi0WS00T23} developed DFedSAM and DFedSAM-MGS, the former utilizing gradient perturbation for local model flattening, and the latter enhancing this with Multiple Gossip Steps for improved model consistency and communication balance. Liu \emph{et al.} \cite{DBLP:conf/infocom/LiuCW22} implemented a method balancing communication efficiency and model consensus through periodic local updates and inter-node communications. Chen \emph{et al.}  \cite{DBLP:conf/icde/ChenXXH23} optimized resource efficiency by limiting gradient pushes to a neighbor subset. Dai \emph{et al.} \cite{DBLP:conf/icml/Dai0H0T22} and Bornstein \emph{et al.} \cite{DBLP:conf/iclr/BornsteinRWBH23} explored various decentralized communication topologies, such as random, ring, and cluster, offering potential solutions for diverse social needs.

Asynchronous FL represents a significant advancement in FL methodologies. This variation has been explored in various studies.
Hagos \emph{et al.} \cite{10226108} ventured beyond centralized FL by investigating scalable asynchronous FL techniques for privacy-preserving real-time surveillance systems, demonstrating its practical applications.
Jang \emph{et al.} \cite{DBLP:conf/aaai/JangL22} introduced AsyncFL, a system where clients upload models based on their capabilities, and the FL server asynchronously updates and broadcasts the global model, showcasing the flexibility of asynchronous operations.
Yang \emph{et al.} \cite{DBLP:conf/icml/YangZK022} developed Anarchic FL, allowing workers to decide their participation schedule and the extent of local computations in each round, emphasizing the autonomy in centralized FL settings.
Bornstein \emph{et al.} \cite{DBLP:conf/iclr/BornsteinRWBH23} proposed SWIFT for decentralized FL, enabling participants to join training asynchronously and broadcast model parameters at their convenience, highlighting the adaptability in decentralized environments.
%


FL plays a pivotal role in ensuring data security and privacy, adhering to strict regulations like the EU/UK General Data Protection Regulation, thereby becoming increasingly relevant in healthcare \cite{DBLP:journals/compsec/TruongSWGG21}. Its application across various healthcare scenarios \cite{10.1145/3533708,DBLP:journals/titb/AkterMLR22, DBLP:journals/cluster/RahmanHMKDRKTB23, DBLP:journals/titb/HeLBWWKY23}, especially including diabetes detection \cite{DBLP:conf/aiccsa/DoloLB22} and prediction \cite{DBLP:journals/bspc/SuHZLJ23, DBLP:series/lncs/GuoYKTCDY20}, and glucose prediction \cite{DBLP:journals/sensors/FalcoCKUKST23}, underscores its significance in protecting healthcare data. Notably, Falco \emph{et al.} \cite{DBLP:journals/sensors/FalcoCKUKST23} introduced a FL framework using an evolutionary algorithm in diabetes management marks a significant stride in privacy-focused glucose prediction. However, existing studies in healthcare often overlook the full potential of FL in terms of generalization, scalability, and asynchronous updates, essential for real-world healthcare implementation. Centralized training frameworks fall short in addressing diverse ``social preferences'' and ``device limitations''. To the best of our knowledge, our proposed approach, GluADFL, is the first work that overcomes these limitations by employing asynchronous decentralized FL aided by various communication graphs for learning population models, thus enhancing the practicality and applicability of FL in healthcare.

\section{Proposed Method}
\subsection{Problem Definition}
\label{sec:pro_def}
Our goal is to develop an asynchronous decentralized FL framework, aimed at enhancing diabetes management for both seen and unseen patients. While this support initially focuses on Blood Glucose Level Prediction (BGLP), it's designed to be adaptable for broader applications such as insulin guidance, dietary recommendations, and other critical aspects of life-cycle management. In this paper, we concentrate specifically on BGLP.

\textbf{BGLP (BG only)}: We define BGLP as the task of predicting future glucose levels $x_{L+H}$, based on a series of historical glucose records $x_{1:L} = \{x_1, ..., x_L\}$, collected through CGM at regular intervals $\delta l$ (e.g., every 5 minutes). The prediction horizon, denoted as $H$, is the future time when the prediction is made (e.g., $H=6$ equals to 30 minutes ahead).

Our approach focuses on using univariate glucose series, denoted as $x_{1:L}$.
This approach stems from the practical challenges in collecting high-quality data via a full suite of wearable devices from all patients \cite{Canali2022} and ensuring their comfort with wearing multiple sensors.
Self-reported data also presents the risk of artifacts and errors in logging \cite{sameera2021human} and could lead to an unpleasant user experience due to the frequent need for manual entry. 
%
%
Thus, relying solely on CGM data emerges as a more convenient and accurate method for future BG predictions. This decision aligns with research underscoring the substantial impact of glucose trajectories on prediction accuracy \cite{DBLP:conf/iscas/ZhuCKZLG23, piao2023blood, prendin2023importance}.
For instance, our previous work \cite{DBLP:conf/iscas/ZhuCKZLG23} has demonstrated that CGM data and timestamps together contribute significantly (93.9\%) to future BG level predictions, with the major contribution attributed to CGM. Meanwhile, there is substantial variability in self-reported events among individuals, which can impair the predictive performance of population models. This issue is particularly crucial for FL that aims to generalize models across diverse populations.

An additional benefit of adopting a univariate approach is the optimization of computational resources and its effective performance, especially in LSTM models \cite{DBLP:conf/ecai/BevanC20, DBLP:journals/npjdm/ZahedaniVMAHARTWHS23}. This not only enhances performance but also significantly reduces the training load on mobile devices, making the process more efficient and feasible for widespread use in patient care.

\subsection{Problem Formulation}
As outlined in section \ref{sec:introduction}, our objective is to utilize FL for developing a population model that addresses the ``cold start'' issue for unseen patients while simultaneously enhancing the prediction accuracy of personalized models for seen patients. The population model is characterized as follows:

\textbf{Population Modeling}: Given a set of patients $\mathcal{N}=\{1,...,n,...,N\}$, where each patient $n$ possesses a private dataset $\mathcal{X}^n$, our goal is to derive a population model $\hat{x}^n_{L+H}=f(x^n_{1:L}; \bw)$ for all $n$ in $\mathcal{N}$, with $\bw$ representing the model's learnable parameters, and $H$ is the prediction horizon.

In contrast to the personalized model $f(x^n_{1:L}; \bw^n)$, which tailors to individual data of the patient $n$, the population model $f(x^n_{1:L}; \bw)$ utilizes non-personalized parameters $\bw$ to capture universal patient patterns \cite{DBLP:conf/ecai/BevanC20}. This approach effectively tackles the ``cold start'' issue for patients not previously encountered in the model \cite{DBLP:conf/embc/MohebbiJHCTJBM20}.
In this work, we adopt the LSTM \cite{DBLP:journals/neco/HochreiterS97} for the population model.
The LSTM framework captures long-term dependencies via its self-gated mechanism, which includes input, forget, and output gates, allowing for effective information flow management. Unlike convolutional methods that lack memory for long-term sequences, LSTM excels in recognizing patterns over extended periods. 
Moreover, this technology strikes an optimal balance by offering the capacity to learn complex temporal patterns with minimal need for extensive hyper-parameter tuning, ensuring both resource conservation and effective performance in constrained environments.
Its blend of computational efficiency, effectiveness, and the presence of well-developed APIs renders it particularly suitable for mobile devices.

Furthermore, we aim to adopt decentralized FL to facilitate increased participation without the scalability constraints associated with a central server. This approach encourages a broader inclusion of participant data in the learning process.

\textbf{Decentralized FL}: In a decentralized FL setup, nodes $\mathcal{N} = \{1, ..., n, ..., N\}$ are interconnected via a communication graph $\mathcal{G}$, mirroring the patient set $\mathcal{N}$. Each patient $n$ corresponds to a node in $\mathcal{G}$, with the graph's edges representing the communication links between patients. As for node $n$, this topology allows it to direct communicate with its neighboring nodes $\mathcal{N}^n$ connected by one-hop links. The decentralized FL framework aims to optimize a global function:
\begin{equation}
    \mathop{\min}\limits_{\bw} \sum_n J(x_{L+H}^n, \bw),
\end{equation}
where the loss function for the BGLP at each node $n$ is defined as:
\begin{equation}
    J(x_{L+H}^n, \bw) = \mathbb{E}_{x^n\sim\mathcal{X}^n}[(x_{L+H}^n - f(x^n_{1:L}; \bw))^2],
    \label{Eqn:loss}
\end{equation}
aiming to minimize the mean square error of all nodes between predicted and actual BG levels, thus improving prediction accuracy in BGLP through collaborative learning without centralized control.

\subsection{Proposed Asynchronous Decentralized FL for Learning a Population Model in BGLP}

Figure \ref{fig:model}a depicts participants (seen patients with T1D who possess sufficient CGM data) as nodes in the communication graph $\mathcal{G}$, representing a network that can adopt ring, cluster, or random topologies as shown in Figure \ref{fig:model}b. This topology choice is based on the participants' communication preferences. CGM devices collect glucose data for processing and training on local devices. These devices engage in GluADFL, an asynchronous decentralized FL process for BGLP, working collectively to create a population model that supports BG management for new users (unseen patients with T1D who do not have enough CGM data).

The GluADFL process entails:
\begin{itemize}
    \item \textbf{Step 1 - Initial Setup}: Node models are initialized with random parameters (Line 3).
    \item \textbf{Step 2 - Broadcasting}: Active nodes share their model parameters with nearby active nodes, influenced by $\mathcal{G}$'s topology (Line 5).
    \item \textbf{Step 3 - Local Model Aggregation}: Each active node update its model by averaging its own model and the received models from neighbors (Lines 7-9).
    \item \textbf{Step 4 - Local Training}: Nodes refine their models using their data based on aggregated parameters (Lines 11-13).
    \item \textbf{Step 5 - Iterative Learning}: Steps 2-4 are repeated asynchronously until achieving convergence or meeting a stop condition. 
    \item \textbf{Step 6 - Population Model Formation}: Final model parameters are aggregated from all nodes to form the population model (Lines 15-16).
\end{itemize}

\begin{algorithm}[tb]
	\caption{GluADFL}\label{Alg:gluadfl}
    \textbf{Input}: Training steps $T$, set of nodes $\mathcal{N}$, max communication batch size $B$, initial model parameters $\{\bw_0^n\}_1^N$, learning rate $\gamma$, and training data $\{\mathcal{X}^n\}_1^N$\\
    \textbf{Output}: Population model $\bw$
    \BlankLine
    
    Initialize node models in $\mathcal{N}$ with $\{\bw_0^n\}_1^N$
    \BlankLine
    \For{$t=1$ \KwTo $T$}{
        \textbf{Broadcast} model parameters from active nodes to their neighbors, depending on the current communication graph
        \BlankLine
        
        \# Model updates post-broadcast\\
        \For{$n=1$ \KwTo $N$}{
            \If{node $n$ is active}{
                Update node $n$'s model with the neighbor node set $\mathcal{N}_t^n$, ensuring $\left|\mathcal{N}_t^n\right|\leq B$: $\hat{\bw}_{t-1}^n \leftarrow \frac{1}{\left|\mathcal{N}_t^n\right| + 1}\left(\sum_{n^\prime\in \mathcal{N}_t^n}\bw_{t-1}^{n^\prime} + \bw_{t-1}^n\right)$
            }
        }
        \BlankLine
        \# Node model updates with local data\\
        \For{$n=1$ \KwTo $N$}{
            \If{node $n$ is active}{
                Update using local data: $\bw^n_t \leftarrow \hat{\bw}_{t-1}^n - \gamma \nabla J(x_{L+H}^n, \bw_{t-1}^n)$
            }
        }
    }
    \BlankLine
    \# Finalize the population model $\bw$\\
    \textbf{Aggregate} model parameters from all nodes\\
    $\bw \leftarrow \frac{1}{N}\sum_n \bw_T^n$
\end{algorithm}

Convergence (in step 5) within the context of GluADFL can be defined as reaching a state where the objective function, as detailed in Equation (\ref{Eqn:loss}), cannot be minimized any further. This indicates that the model has reached its optimal performance, and further iterations do not yield significant improvements in prediction accuracy.
Additional stopping criteria for the algorithm can include specific thresholds for the objective function. This allows for flexibility in determining when the learning process has reached a satisfactory level of performance or optimization, providing a practical mechanism for stopping the algorithm under predefined conditions.

The algorithm enables the direct application of the population model for unseen patients, offering immediate benefits in BG management. Seen patients, who have contributed their data, have the flexibility to further refine this population model with their own data, utilizing Equation (\ref{Eqn:loss}) and adjusting the learning rate $\gamma$ for personalization. This customization process enhances the model's relevance to the individual's unique data patterns. Moreover, seen patients might also directly use the population model, as it often surpasses solely personalized models that lack population-derived insights, providing a robust starting point for individual adaptations.

The GluADFL framework facilitates asynchronous communication \cite{DBLP:conf/iclr/BornsteinRWBH23}, allowing nodes to update their models or pause without waiting for others, enhancing flexibility and efficiency. Nodes can rejoin the learning process when ready, ensuring that each participant contributes at their own pace. This wait-free mechanism supports continuous local model improvements and collective learning without the need for synchronous updates, making the learning process more adaptable to individual and network variabilities.

In the GluADFL framework, we explore three distinct communication graphs to facilitate model training and parameter sharing:
\begin{itemize}
\item \textbf{Random} \cite{DBLP:conf/icml/Dai0H0T22}: As illustrated in Figure \ref{fig:model}a, active nodes randomly establish connections with up to $B$ other active nodes for exchanging model parameters. This topology introduces a flexible and dynamically changing network structure at each training step.
\item \textbf{Cluster} \cite{DBLP:conf/iclr/BornsteinRWBH23}: Initially, nodes are organized into clusters, creating a ring-like structure where each cluster is fully connected internally and linked to others through specific nodes. This setup, shown in Figure \ref{fig:model}b, promotes efficient intra- and inter-cluster communication, with the structure remaining constant throughout training.
\item \textbf{Ring} \cite{DBLP:conf/iclr/BornsteinRWBH23}: Nodes are arranged in a circular pattern, enabling sequential parameter sharing along the ring. This ensures a consistent, cyclic flow of information among all active participants, depicted in Figure \ref{fig:model}b, with the communication pattern fixed during the training.
\end{itemize}

The choice of communication graphs in GluADFL can cater to diverse social preferences and privacy concerns of participants. The random topology fits scenarios where participants are open to broader interaction beyond their immediate social circles. The cluster topology is ideal for those preferring interactions within known groups, resembling social circles.
For utmost privacy, the ring topology allows communication exclusively between two closely connected individuals, mirroring the preference to limit data (model parameters) sharing to a minimum. Thus, these topologies offer customizable communication settings to meet varied social and privacy needs.

These communication graphs enhance the distributed nature of communication and decrease dependency on a single master node, characteristic of the star topology in FedAvg \cite{DBLP:conf/aistats/McMahanMRHA17}, as illustrated in Figure \ref{fig:model}b. By facilitating efficient, decentralized exchanges among participants, these structures foster collaboration and accelerate the learning process for BGLP, offering a more resilient and scalable approach to FL.

\section{Experiments and Results}
\subsection{Datasets}
All the results are based on the prediction of future BG levels for patients with T1D using OhioT1DM \cite{DBLP:conf/ecai/MarlingB20}, ABC4D (\cite{Reddy2016,DBLP:journals/iotj/ZhuKDHLG23}, NCT02053051), CTR3 (NCT02137512) and REPLACE-BG (\cite{Aleppo2017ReplaceBG}, NCT02258373). 
The statistics for these four datasets are presented in Table \ref{table:statis}, highlighting their diversity. For example, the ABC4D dataset delivered insulin exclusively using pens (Novo Nordisk Echo), whereas other datasets leveraged pumps. This potentially makes the BG variability the greatest in ABC4D, resulting in a challenging BG prediction.
Each dataset, excluding OhioT1DM, is divided into training ($60\%$), validation ($20\%$) and testing data ($20\%$) by time per person. As for OhioT1DM, it has been originally divided into training and testing data. We make the last $20\%$ of its training data as the validation data. As mentioned in section \ref{sec:pro_def}, only BG time series is considered. Hence, we leveraged Z-Score normalization to standardize the BG levels within each dataset using the mean and standard deviation of training data, then all missing values are replaced with zero \cite{DBLP:conf/ecai/BevanC20}.

\begin{table*}[tb]
\centering
\caption{The statistics of four datasets}
\resizebox{1.0\textwidth}{!}{
\begin{tabular}{|c|c|c|c|c|}
\hline
Demographic              & OhioT1DM      & ABC4D         & CTR3          & REPLACE-BG    \\ \hline 
CGM                      & Medtronic Enlite &  Dexcom G5 & Dexcom G4 & Dexcom G4 \\ \hline
Insulin Pump or Pen                    & MiniMed 530G/630G & Novo Nordisk Echo & Roche Accu-Chek & MiniMed 530G, OmniPod, etc. \\ \hline
No. of Participants      & 12            & 25            & 30            & 226           \\ \hline
No. of Days              & 54(2)         & 168(14)       & 163(67)       & 251(39)       \\ \hline
No. of CGM Records per Participant       & 13871(1015)   & 43259(5460)   & 43421(18309)  & 66153(10701)  \\ \hline
Mean of CGM Data (mg/dL) & 159.35(16.34) & 156.66(24.24) & 151.37(13.34) & 160.69(21.18) \\ \hline
SD of CGM Data (mg/dL)   & 58.11(6.15)   & 60.52(14.47)  & 55.29(8.24)   & 60.33(11.65)  \\ \hline
Time In Ringe ($\%$)                & 63.54(9.70)   & 62.54(15.58)  & 69.92(7.95)   & 63.10(12.18)  \\ \hline
Time Below Ringe ($\%$)                & 3.30(2.25)    & 6.01(4.13)    & 3.53(2.11)    & 3.78(2.51)    \\ \hline
Time Above Ringe ($\%$)                & 33.15(10.71)  & 31.45(15.65)  & 26.54(8.57)   & 33.13(12.93)  \\ \hline
Coefficient of Variation ($\%$)                 & 36.63(3.70)   & 38.40(6.22)   & 36.44(3.90)   & 37.45(4.70)   \\ \hline
Low Blood Glucose Index                     & 0.88(0.48)    & 1.73(1.02)    & 0.97(0.48)    & 1.00(0.58)    \\ \hline
High Blood Glucose Index                     & 7.15(2.45)    & 7.26(3.85)    & 5.89(1.93)    & 7.57(3.42)    \\ \hline
\end{tabular}
}
\label{table:statis}
\end{table*}

\subsection{Ethics and Data availability}

All datasets can be accessed publicly apart from ABC4D, which can be accessed via authorised procedures by contacting the project manager and the corresponding author. The ABC4D studies were conducted under protocol (13/LO/0264) approved by the London - Chelsea Research Ethics Committee in 2013.

\subsection{Metrics}
In this study, the metrics used to evaluate the performance of the BGLP includes root mean square error (RMSE), mean absolute relative difference (MARD), mean absolute error (MAE), glucose-specific RMSE (gRMSE, \cite{DBLP:journals/tbe/FaveroFC12, DBLP:journals/npjdm/ZhuULHOG22}) and time lag. 
RMSE (mg/dL) measures the square root of the average of the squares of the errors, as follows ($I$ is the total number of testing examples): 
\begin{equation}
    RMSE = \sqrt{\frac{1}{I} \sum_i({x}_{L+H, i} - \hat{x}_{L+H, i})^2}.
\end{equation}

MARD (\%) averages the absolute differences between predicted and actual values, expressed as a percentage of the actual values. It is used to assess the relative accuracy of the predictions. 
\begin{equation}
    MARD = \frac{1}{I} \sum_i\frac{|{x}_{L+H, i} - \hat{x}_{L+H, i}|}{{x}_{L+H, i}} \times 100\%.
\end{equation}

MAE (mg/dL) measures the average of the absolute differences between the predicted values and the actual values. Without squaring the errors before averaging allows it to be less sensitive to large errors than RMSE. 
\begin{equation}
    MAE = \frac{1}{I} \sum_i \left|{x}_{L+H, i} - \hat{x}_{L+H, i}\right|.
\end{equation}

gRMSE (mg/dL) is a variation of RMSE, evaluating the prediction error specifically within certain ranges of BG levels. It allows for a more detailed assessment of model performance across different glucose level ranges, which is crucial in diabetes management where different ranges (e.g., hypoglycemia, hyperglycemia) have different clinical implications.
\begin{equation}
    gRMSE = \sqrt{\frac{1}{I} \sum_i P({x}_{L+H, i}, \hat{x}_{L+H, i})({x}_{L+H, i} - \hat{x}_{L+H, i})^2},   
\end{equation}
where $P({x}_{L+H, i}, \hat{x}_{L+H, i})$ penalizes overestimation in hypoglycemia and underestimation in hyperglycemia, and more details are in \cite{DBLP:journals/tbe/FaveroFC12}. 


In BGLP, time lag (minutes) quantifies the temporal discrepancy between actual BG level changes and when these changes are detected by the model's predictions. This concept, as detailed by Cohen \cite{cohen1995time} through cross-correlation analysis, and further applied in BGLP contexts \cite{DBLP:journals/titb/GaniGLWVR10, DBLP:journals/titb/LiLZHG20}, highlights the critical challenge of timely and accurate BG prediction, emphasizing the importance of minimizing lag for effective diabetes management.

In summary, we pursue lower values of all these metrics, meaning better prediction models for BGLP.

\subsection{Baselines}
\label{sec:baselines}
In our experiments, several baselines were compared with the proposed model. First of all, five kinds of population models were gotten by traditional supervised learning through mixing the training data of all patients within each dataset as follows:
\begin{itemize}
    \item \textbf{LR}:it is a Linear Regression (LR) model that predicts values based on linear relationships between input features and target variable.
    \item \textbf{XGBoost} \cite{DBLP:conf/kdd/ChenG16}: it is based on gradient boosting, leveraging decision trees and optimizing model performance through sequential learning.
    \item \textbf{LSTM}: a single layer of LSTM whose self-gated mechanisms can model long short-term patterns of time series.
    \item \textbf{N-BEATS} \cite{DBLP:conf/iclr/OreshkinCCB20}: it is a Neural Basis Expansion Analysis for interpretable Time Series (N-BEATS) forcasting through leveraging a stack of fully-connected layers and backcasting.
   \item \textbf{NHiTS} \cite{DBLP:conf/aaai/ChalluOORCD23}: it is a Neural Hierarchical Interpolation for Time Series (NHiTS), leveraging hierarchical interpolation and pooling techniques. Compared with N-BEATS, it improves time series predictions by specializing in different frequencies and reducing computational complexity.
\end{itemize}

Furthermore, we introduced meta-learning methods and centralized FL to our baselines for comparisons.
\begin{itemize}
    \item \textbf{MAML} \cite{DBLP:conf/icml/FinnAL17}: an LSTM model trained by Model-Agnostic Meta-Learning (MAML).
MAML is to find a set of initial parameters that can quickly adapt to new tasks through few gradient updates.
    \item \textbf{MetaSGD} \cite{DBLP:journals/corr/LiZCL17}: an LSTM model trained by Meta-learner acting like Stochastic Gradient Descent (MetaSGD). MetaSGD extends MAML by introducing a learnable learning rate for each parameter in the model, providing more flexibility and potentially to various tasks.
    \item \textbf{FedAvg} \cite{DBLP:conf/aistats/McMahanMRHA17}: an LSTM model trained by using centralized Federated Averaging (FedAvg) method. It has a centralized server to periodically broadcast latest model parameters to local devices and average the local model parameters which are trained separately by local data.
\end{itemize}

Then, our proposed method with three different communication graphs are denoted as ``GluADFL(Ring)'', ``GluADFL(Cluster)'' and ``GluADFL(Random)'' . 

We set the historical BG series length as $L = 12$, i.e., 2 hours, predicting the BG level in 30 minutes ($H=6$). 
Except for LR, each method was trained four times with different random seeds. We leverage grid search for selecting the hyperparameters of XGBoost.
%
We implemented all deep learning methods using PyTorch 1.11.0, based on \cite{learn2learn2019,
sklearn_api} and PyTorch Forcasting\footnote{https://github.com/jdb78/pytorch-forecasting}, and executed them on an NVIDIA RTX 3090 Ti. For deep learning methods, the learning rate was found in $\{10^{-3}, 10^{-4}, 10^{-5}\}$ and the hidden state size was found in $\{128, 256, 512\}$. The best hyperparameters were selected based on validation data performance. All experiments and results can be accessible and reproducible: ``https://github.com/ChengzhePiao/coldstartbglp''.

\subsection{Results}


To assess the cross-prediction capabilities of population models for unseen patients, we compare the performance of population LSTM models trained via the ``GluADFL (Random)'' approach (Table \ref{table:gen_gluadfl}) and traditional supervised learning (Table \ref{table:gen_mixing}) across four distinct datasets. Performance is evaluated using testing data from all datasets, with specific interest in the off-diagonal cells (non-bold) in Tables \ref{table:gen_gluadfl} and \ref{table:gen_mixing}, which represent predictions for unseen patients. 
%
In the analysis, differences in performance between seen and unseen patients are categorized by color: differences below $0.30$ or between $(0.3, 0.5]$ are marked in red or blue, respectively. In Table \ref{table:gen_gluadfl}, $78\%$ of metrics for unseen patients show minor discrepancies ($\leq 0.50$) compared to seen patients. Excluding models trained with the OhioT1DM dataset, this percentage rises to $87\%$, suggesting a high level of predictive consistency across different patient groups. This indicates that models trained through GluADFL generally maintain uniform accuracy. However, models trained specifically with OhioT1DM data display weaker generalization, mainly due to the dataset's smaller size. Similarly, in Table \ref{table:gen_mixing}, $80\%$ of metrics for unseen patients are closely matched ($\leq 0.50$) with those of seen patients, increasing to $88\%$ when excluding OhioT1DM-trained models, showcasing the inherent generalization capability of the population models by tradition supervised learning across patient data. Incorporating FL preserves this capability with only a slight reduction.

\begin{figure*}[tb]
\centering
\includegraphics[width=\linewidth]{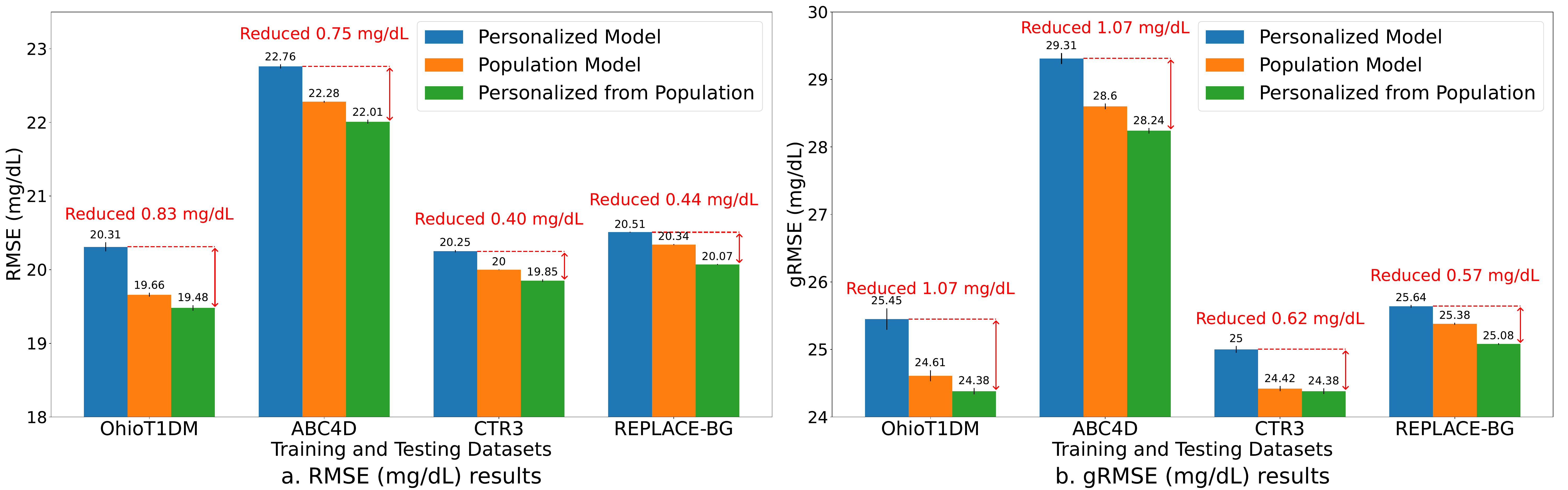}
\caption{Evaluation of ``Personalized Model'', ``Population Model'', and ``Personalized from Population'' across four datasets. The ``Personalized Model'' involves individual models for each patient, originally randomly initialized and trained with the patient's own data. The ``Population Model'' is derived from the GluADFL framework using a random topology and trained by mixing patients' data.  ``Personalized from Population'' approach refines the GluADFL-based population model by integrating individual patient data to create a customized model, offering a hybrid approach that leverages both broad population insights and specific patient data.}
\label{fig:fine_tune}
\end{figure*}

In Figure \ref{fig:fine_tune}, we examine ``Personalized Model'', ``Population Model'', and ``Personalized from Population'' across four datasets, focusing on the advantage for seen patients participating in FL training. A ``Personalized Model'' starts with a randomly initialized model for each participant, individually tailored with their data. The ``Population Model'' is derived from ``GluADFL''. ``Personalized from Population'' allows further refining the population model with personal data, enhancing it into a personalized model. In terms of RMSE, ``Personalized from Population'' outperforms ``Personalized Model'' by $0.83$, $0.75$, $0.40$, and $0.44$ mg/dL, and in terms of gRMSE, by $1.07$, $1.07$, $0.62$, and $0.57$ mg/dL across the datasets, indicating significant benefits for seen patients from incorporating population features through FL training.

\begin{table}[tb]
\centering
\caption{Generalization of population models trained by GluADFL}
\begin{threeparttable}
\resizebox{0.48\textwidth}{!}{
\begin{tabular}{|c|c||c||c|c|c||}
\hline
\multirow{2}{*}{Training Data} & \multirow{2}{*}{Metric} & \multicolumn{4}{c||}{Testing Data}                                                                                                                        \\ \cline{3-6} 
                               &                         & \multicolumn{1}{c||}{OhioT1DM}                 & \multicolumn{1}{c||}{ABC4D}                & \multicolumn{1}{c||}{CTR3}                 & REPLACE-BG           \\ \hline
\multirow{5}{*}{OhioT1DM}          & RMSE                    & \multicolumn{1}{c||}{\textbf{19.66(2.52)}} & \multicolumn{1}{c||}{23.61(4.44)}          & \multicolumn{1}{c||}{{\color{blue}20.49(3.23)}}          & {\color{blue}20.73(3.45)}          \\
                               & MARD                    & \multicolumn{1}{c||}{\textbf{9.43(1.62)}}  & \multicolumn{1}{c||}{12.97(3.73)}          & \multicolumn{1}{c||}{{\color{red}10.38(1.50)}}          & {\color{red}10.71(1.95)}          \\
                               & MAE                     & \multicolumn{1}{c||}{\textbf{13.80(1.61)}} & \multicolumn{1}{c||}{16.25(2.42)}          & \multicolumn{1}{c||}{{\color{blue}14.41(2.24)}}          & {\color{blue}14.75(2.45)}          \\
                               & gRMSE                   & \multicolumn{1}{c||}{\textbf{24.61(3.35)}} & \multicolumn{1}{c||}{30.65(7.00)}          & \multicolumn{1}{c||}{25.15(4.27)}          & 25.98(4.67)          \\
                               & Time Lag                & \multicolumn{1}{c||}{\textbf{5.22(4.48)}}  & \multicolumn{1}{c||}{{\color{red}6.54(5.05)}}           & \multicolumn{1}{c||}{10.31(3.97)}          & {\color{red}9.41(3.99)}           \\ \hline
\multirow{5}{*}{ABC4D}         & RMSE                    & \multicolumn{1}{c||}{{\color{red}19.74(2.56)}}          & \multicolumn{1}{c||}{\textbf{22.28(4.03)}} & \multicolumn{1}{c||}{{\color{red}20.08(3.21)}}          & {\color{red}20.21(3.22)}          \\
                               & MARD                    & \multicolumn{1}{c||}{{\color{red}9.45(1.61)}}           & \multicolumn{1}{c||}{\textbf{12.16(3.22)}} & \multicolumn{1}{c||}{{\color{red}10.13(1.51)}}          & {\color{red}10.43(1.85)}          \\
                               & MAE                     & \multicolumn{1}{c||}{{\color{red}13.88(1.59)}}          & \multicolumn{1}{c||}{\textbf{15.41(2.05)}} & \multicolumn{1}{c||}{{\color{red}14.11(2.23)}}          & {\color{red}14.43(2.34)}          \\
                               & gRMSE                   & \multicolumn{1}{c||}{{\color{red}24.69(3.26)}}          & \multicolumn{1}{c||}{\textbf{28.60(6.34)}} & \multicolumn{1}{c||}{{\color{red}24.57(4.20)}}          & {\color{red}25.28(4.32)}          \\
                               & Time Lag                & \multicolumn{1}{c||}{{\color{red}5.36(4.17)}}           & \multicolumn{1}{c||}{\textbf{6.69(4.90)}}  & \multicolumn{1}{c||}{10.37(4.01)}          & {\color{red}9.51(3.87)}           \\ \hline
\multirow{5}{*}{CTR3}          & RMSE                    & \multicolumn{1}{c||}{{\color{red}19.76(2.61)}}          & \multicolumn{1}{c||}{22.79(4.19)}          & \multicolumn{1}{c||}{\textbf{20.00(3.13)}} & {\color{red}20.28(3.27)}          \\
                               & MARD                    & \multicolumn{1}{c||}{{\color{red}9.45(1.62)}}           & \multicolumn{1}{c||}{{\color{blue}12.55(3.54)}}          & \multicolumn{1}{c||}{\textbf{10.09(1.47)}} & {\color{red}10.49(1.87)}          \\
                               & MAE                     & \multicolumn{1}{c||}{{\color{red}13.83(1.57)}}          & \multicolumn{1}{c||}{{\color{blue}15.73(2.19)}}          & \multicolumn{1}{c||}{\textbf{14.03(2.17)}} & {\color{red}14.44(2.34)}          \\
                               & gRMSE                   & \multicolumn{1}{c||}{{\color{red}24.71(3.42)}}          & \multicolumn{1}{c||}{29.46(6.66)}          & \multicolumn{1}{c||}{\textbf{24.42(4.07)}} & {\color{red}25.36(4.41)}          \\
                               & Time Lag                & \multicolumn{1}{c||}{{\color{red}5.01(3.76)}}           & \multicolumn{1}{c||}{{\color{red}6.35(4.81)}}           & \multicolumn{1}{c||}{\textbf{9.76(3.79)}}  & {\color{red}9.02(3.79)}
                               \\ \hline
\multirow{5}{*}{REPLACE-BG}    & RMSE                    & \multicolumn{1}{c||}{{\color{red}19.75(2.61)}}          & \multicolumn{1}{c||}{22.97(4.26)}          & \multicolumn{1}{c||}{{\color{red}20.17(3.20)}}          & \textbf{20.34(3.32)} \\
                               & MARD                    & \multicolumn{1}{c||}{{\color{red}9.38(1.61)}}           & \multicolumn{1}{c||}{{\color{blue}12.54(3.46)}}          & \multicolumn{1}{c||}{{\color{red}10.17(1.50)}}          & \textbf{10.45(1.88)} \\
                               & MAE                     & \multicolumn{1}{c||}{{\color{red}13.80(1.59)}}          & \multicolumn{1}{c||}{{\color{blue}15.77(2.19)}}          & \multicolumn{1}{c||}{{\color{red}14.14(2.21)}}          & \textbf{14.44(2.36)} \\
                               & gRMSE                   & \multicolumn{1}{c||}{{\color{red}24.62(3.40)}}          & \multicolumn{1}{c||}{29.68(6.74)}          & \multicolumn{1}{c||}{{\color{red}24.61(4.15)}}          & \textbf{25.38(4.47)} \\
                               & Time Lag                & \multicolumn{1}{c||}{{\color{blue}5.56(4.46)}}           & \multicolumn{1}{c||}{{\color{red}6.64(4.95)}}           & \multicolumn{1}{c||}{10.38(4.10)}          & \textbf{9.46(3.93)}  \\ \hline
\end{tabular}
}
    \begin{tablenotes}
   
    \item {\fontsize{5.5}{1.5}\selectfont \textbf{Bold}: results for seen patients (diagonal);  \par}
    \item {\fontsize{5.5}{1.5}\selectfont Un-Bold: results for unseen patients (non-diagonal);  \par}
    \item {\fontsize{5.5}{1.5}\selectfont {\color{red} Red}: differences between seen patients and unseen patients are below $0.30$ for each column;  \par}
    \item {\fontsize{5.5}{1.5}\selectfont {\color{blue} Blue}: differences between seen patients and unseen patients are between $(0.3, 0.5]$ for each column.  \par}
    
    \end{tablenotes}
  \end{threeparttable}
\label{table:gen_gluadfl}
\end{table}

\begin{table}[tb]
\centering
\caption{Generalization of population models trained by mixing data}
\begin{threeparttable}
\resizebox{0.48\textwidth}{!}{
\begin{tabular}{|c|c||c||c|c|c||}
\hline
\multirow{2}{*}{Training Data} & \multirow{2}{*}{Metric} & \multicolumn{4}{c||}{Testing Data}                                                                                                                        \\ \cline{3-6} 
                               &                         & \multicolumn{1}{c||}{OhioT1DM}             & \multicolumn{1}{c||}{ABC4D}                & \multicolumn{1}{c||}{CTR3}                 & REPLACE-BG           \\ \hline
\multirow{5}{*}{OhioT1DM}      & RMSE                    & \multicolumn{1}{c||}{\textbf{19.65(2.48)}} & \multicolumn{1}{c||}{23.59(4.41)}          & \multicolumn{1}{c||}{20.56(3.28)}          & {\color{blue}20.76(3.45)}         \\
                               & MARD                    & \multicolumn{1}{c||}{\textbf{9.53(1.65)}}  & \multicolumn{1}{c||}{{\color{blue}12.93(3.58)}}          & \multicolumn{1}{c||}{{\color{red}10.46(1.51)}}          & {\color{blue}10.81(1.95)}          \\
                               & MAE                     & \multicolumn{1}{c||}{\textbf{13.89(1.60)}} & \multicolumn{1}{c||}{16.28(2.38)}          & \multicolumn{1}{c||}{{\color{blue}14.49(2.25)}}          & {\color{blue}14.86(2.46)}          \\
                               & gRMSE                   & \multicolumn{1}{c||}{\textbf{24.65(3.26)}} & \multicolumn{1}{c||}{30.60(6.96)}          & \multicolumn{1}{c||}{25.28(4.34)}          & 26.05(4.68)          \\
                               & Time Lag                & \multicolumn{1}{c||}{\textbf{5.11(4.38)}}  & \multicolumn{1}{c||}{{\color{red}6.60(5.13)}}           & \multicolumn{1}{c||}{10.34(4.07)}          & {\color{red}9.43(3.99)}           \\ \hline
\multirow{5}{*}{ABC4D}         & RMSE                    & \multicolumn{1}{c||}{{\color{red}19.87(2.58)}}          & \multicolumn{1}{c||}{\textbf{22.42(4.06)}} & \multicolumn{1}{c||}{{\color{red}20.22(3.13)}}          & {\color{red}20.44(3.23)}          \\
                               & MARD                    & \multicolumn{1}{c||}{{\color{red}9.79(1.71)}}           & \multicolumn{1}{c||}{\textbf{12.51(3.33)}} & \multicolumn{1}{c||}{{\color{red}10.44(1.51)}}          & {\color{blue}10.85(1.94)}          \\
                               & MAE                     & \multicolumn{1}{c||}{{\color{red}14.15(1.64)}}          & \multicolumn{1}{c||}{\textbf{15.69(2.10)}} & \multicolumn{1}{c||}{{\color{red}14.38(2.21)}}          & {\color{red}14.76(2.37)}          \\
                               & gRMSE                   & \multicolumn{1}{c||}{{\color{red}24.83(3.29)}}          & \multicolumn{1}{c||}{\textbf{28.71(6.37)}} & \multicolumn{1}{c||}{{\color{red}24.72(4.07)}}          & {\color{blue}25.55(4.32)}          \\
                               & Time Lag                & \multicolumn{1}{c||}{{\color{blue}5.44(4.20)}}           & \multicolumn{1}{c||}{\textbf{6.77(4.97)}}  & \multicolumn{1}{c||}{10.29(3.85)}          & {\color{red}9.48(3.87)}           \\ \hline
\multirow{5}{*}{CTR3}          & RMSE                    & \multicolumn{1}{c||}{{\color{blue}20.06(2.57)}}          & \multicolumn{1}{c||}{{\color{blue}22.87(4.20)}}          & \multicolumn{1}{c||}{\textbf{20.02(3.06)}} & {\color{red}20.54(3.28)}          \\
                               & MARD                    & \multicolumn{1}{c||}{{\color{red}9.79(1.65)}}           & \multicolumn{1}{c||}{{\color{blue}12.83(3.59)}}          & \multicolumn{1}{c||}{\textbf{10.24(1.46)}} & {\color{blue}10.85(1.93)}          \\
                               & MAE                     & \multicolumn{1}{c||}{{\color{blue}14.24(1.56)}}          & \multicolumn{1}{c||}{{\color{blue}16.01(2.27}}          & \multicolumn{1}{c||}{\textbf{14.17(2.14)}} & {\color{blue}14.83(2.40)}          \\
                               & gRMSE                   & \multicolumn{1}{c||}{25.39(3.37)}          & \multicolumn{1}{c||}{29.72(6.65)}          & \multicolumn{1}{c||}{\textbf{24.70(4.01)}} & 25.97(4.46)          \\
                               & Time Lag                & \multicolumn{1}{c||}{{\color{red}4.85(3.92)}}           & \multicolumn{1}{c||}{{\color{red}6.14(4.71)}}           & \multicolumn{1}{c||}{\textbf{9.21(3.66)}}  & {\color{red}8.55(3.65)}           \\ \hline
\multirow{5}{*}{REPLACE-BG}    & RMSE                    & \multicolumn{1}{c||}{{\color{red}19.75(2.59)}}          & \multicolumn{1}{c||}{{\color{red}22.70(4.14)}}          & \multicolumn{1}{c||}{{\color{red}20.17(3.19)}}          & \textbf{20.31(3.27)} \\
                               & MARD                    & \multicolumn{1}{c||}{{\color{red}9.47(1.65)}}           & \multicolumn{1}{c||}{{\color{red}12.45(3.33)}}          & \multicolumn{1}{c||}{{\color{red}10.22(1.50)}}          & \textbf{10.50(1.88)} \\
                               & MAE                     & \multicolumn{1}{c||}{{\color{red}13.88(1.62)}}          & \multicolumn{1}{c||}{{\color{red}15.69(2.12)}}          & \multicolumn{1}{c||}{{\color{red}14.21(2.22)}}          & \textbf{14.49(2.35)} \\
                               & gRMSE                   & \multicolumn{1}{c||}{{\color{red}24.56(3.35)}}          & \multicolumn{1}{c||}{{\color{blue}29.15(6.54)}}          & \multicolumn{1}{c||}{{\color{red}24.55(4.14)}}          & \textbf{25.24(4.37)} \\
                               & Time Lag                & \multicolumn{1}{c||}{{\color{blue}5.45(4.27)}}           & \multicolumn{1}{c||}{{\color{red}6.71(4.97)}}           & \multicolumn{1}{c||}{10.38(3.98)}          & \textbf{9.50(3.93)}  \\ \hline
\end{tabular}
}
    \begin{tablenotes}
   
    \item {\fontsize{5.5}{1.5}\selectfont \textbf{Bold}: results for seen patients (diagonal);  \par}
    \item {\fontsize{5.5}{1.5}\selectfont Un-Bold: results for unseen patients (non-diagonal);  \par}
    \item {\fontsize{5.5}{1.5}\selectfont {\color{red} Red}: differences between seen patients and unseen patients are below $0.30$ for each column;  \par}
    \item {\fontsize{5.5}{1.5}\selectfont {\color{blue} Blue}: differences between seen patients and unseen patients are between $(0.3, 0.5]$ for each column.  \par}
    
    \end{tablenotes}
  \end{threeparttable}
\label{table:gen_mixing}
\end{table}

\begin{figure*}[tb]
\centering
\includegraphics[width=\linewidth]{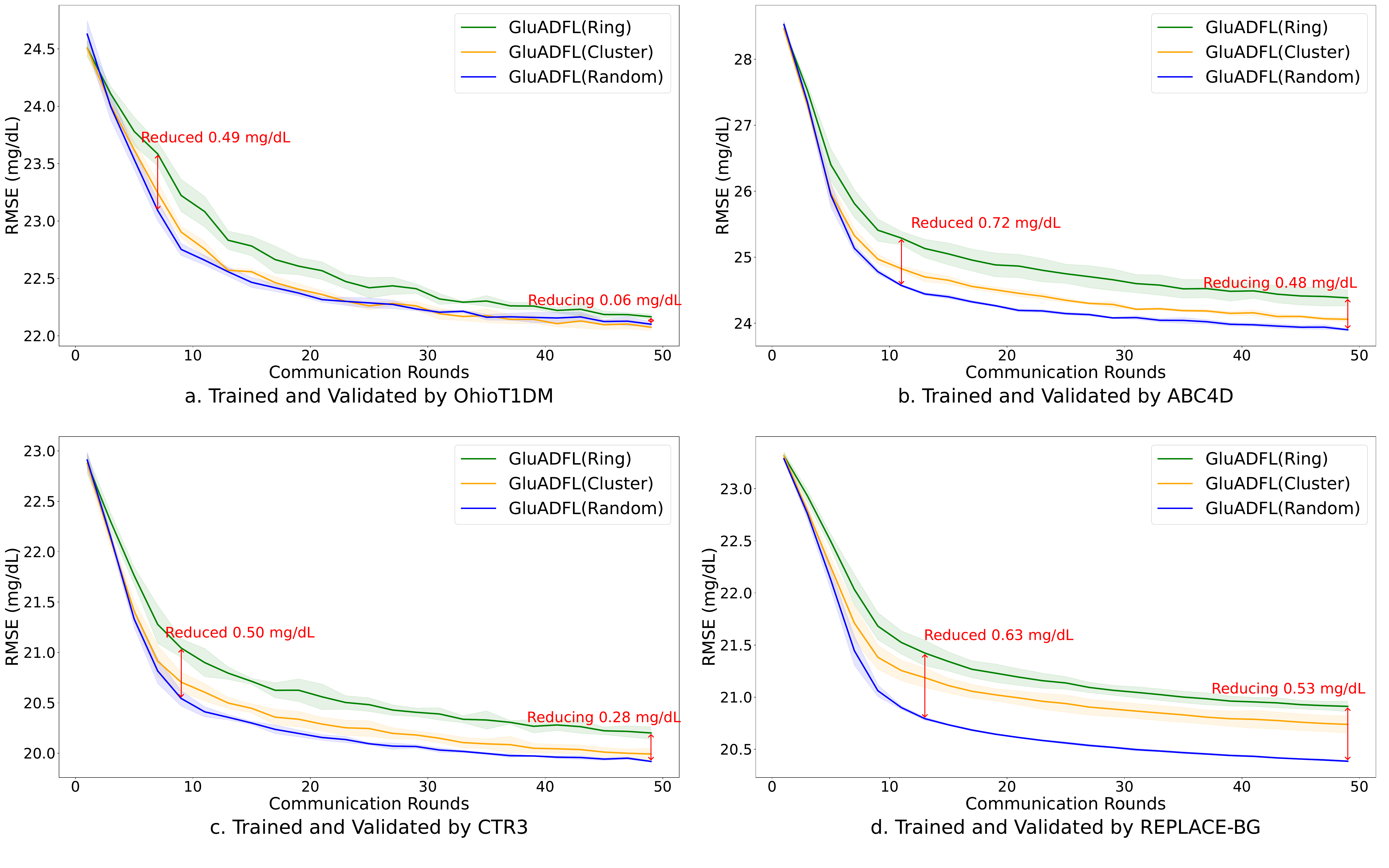}
\caption{The convergence of population models trained by GluADFL with different communication graphs on four datasets. }
\label{fig:rmse_converg}
\end{figure*}

\begin{figure*}[tb]
\centering
\includegraphics[width=\linewidth]{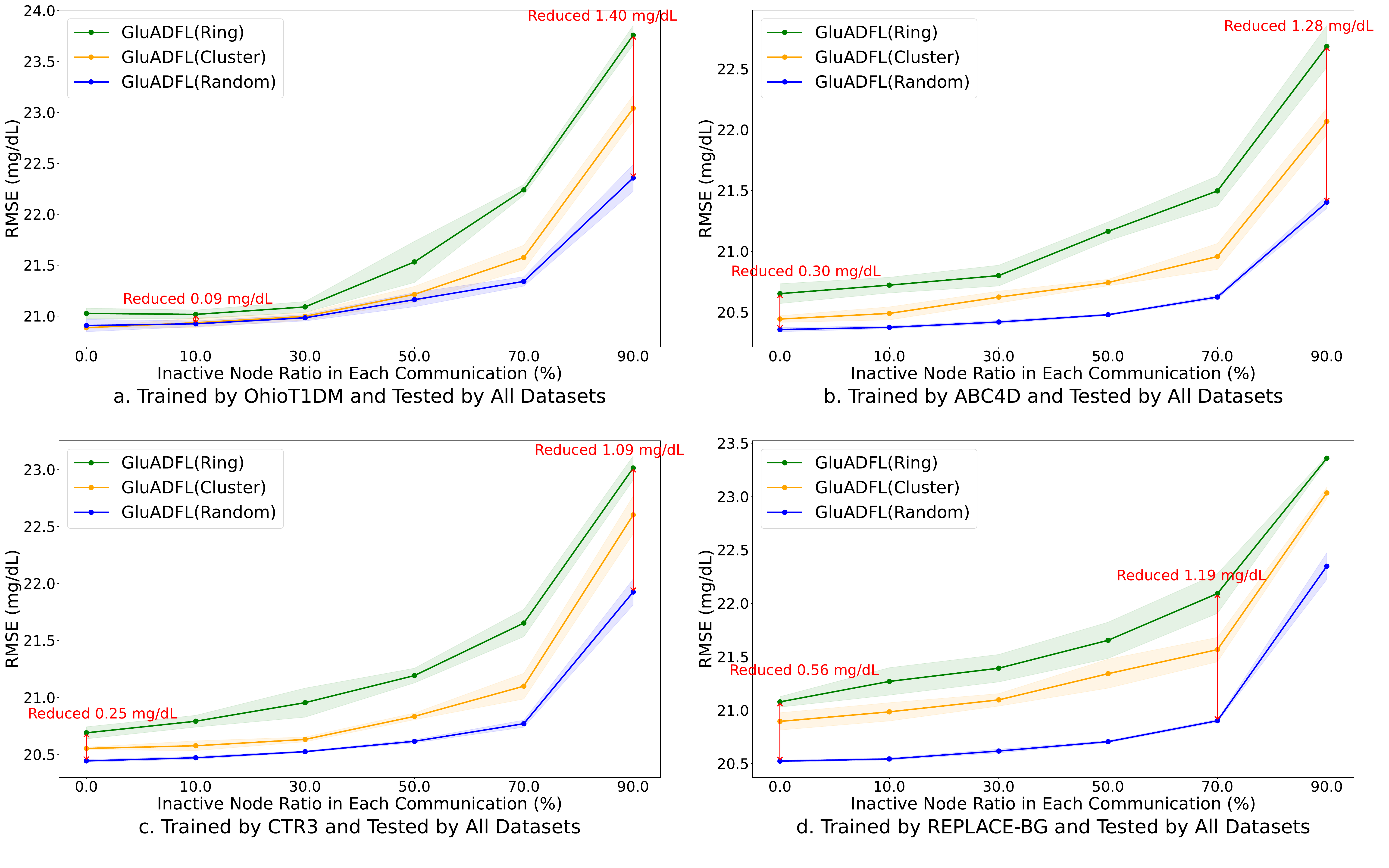}
\caption{The performance of GluADFL with different communication topology when changing inactive node ratio in each communication.}
\label{fig:rmse_inactive}
\end{figure*}

In our analysis, we investigate the impact of different typologies of GluADFL on the training process, as depicted in Figure \ref{fig:rmse_converg}. We observe that, towards the end of the communication rounds, the performance of ``GluADFL(Random)'' surpasses that of ``GluADFL(Ring)'' across various datasets. Specifically, improvements in RMSE for ``GluADFL(Random)'' compared to ``GluADFL(Ring)'' are noted as $0.06$, $0.48$, $0.28$, and $0.53$ mg/dL for OhioT1DM, ABC4D, CTR3, and REPLACE-BG datasets, respectively. The performance of ``GluADFL(Cluster)'' is found to be intermediate between the two.

Furthermore, we assess the influence of varying inactive participant proportions during the training phases on the overall performance across different topologies, as shown in Figure \ref{fig:rmse_inactive}. The evaluation highlight that with a $90\%$ inactive node ratio, ``GluADFL(Random)'' demonstrates a reduction in RMSE by $1.40$, $1.28$, and $1.09$ mg/dL on the OhioT1DM, ABC4D, and CTR3 datasets, respectively, when compared with ``GluADFL(Ring)''. Additionally, at a $70\%$ inactive node ratio, ``GluADFL(Random)'' shows a reduction in RMSE by $1.19$ mg/dL on the REPLACE-BG dataset compared with ``GluADFL(Ring)''.
Overall, ``GluADFL(Random)'' achieve an average reduction in RMSE of $0.50$, $0.64$, $0.59$, and $0.87$ mg/dL across the OhioT1DM, ABC4D, CTR3, and REPLACE-BG datasets, respectively, indicating a superior performance relative to ``GluADFL(Ring)'', with ``GluADFL(Cluster)'' maintaining a performance level that is between the two.

We detail the comparative results of prediction performance across diverse methods, as captured in Table \ref{table:compare_seen_unseen}. This table illustrates the efficacy of each method when trained on a specified dataset and subsequently tested on both seen (identical dataset) and unseen (alternative datasets) patient data. This comparison aims to evaluate the generalizability of each model comprehensively. The performance of the LSTM model, trained through supervised learning, is highlighted in bold, establishing a benchmark for comparison. Discrepancies from this benchmark are indicated visually in red for differences $\leq 30$ and in blue for differences within the $(30, 50]$ range.
\begin{itemize}
    \item Among the evaluated population models LR, XGBoost, LSTM, N-BEATS and NHiTS, the LSTM model outperforms in terms of accuracy for both seen and unseen patient data. LR and XGBoost are the least effective, while N-BEATS and NHiTS lag slightly behind LSTM, with $80\%$ and $75\%$ of their results showing only a $0.3$ disparity with LSTM for seen and unseen patients, respectively.
    \item In the context of meta-learning and FL versus traditional supervised learning, meta-learning methods do not significantly surpass the benchmark. FL-based methodologies, particularly FedAVG and ``GluADFL(Random)'', achieve results comparable to supervised learning, with $95\%$ of their outcomes differing from LSTM's by $0.3$ across both seen and unseen patient data scenarios. ``GluADFL(Ring)'' shows relatively lower effectiveness, with $60\%$ and $65\%$ of its results showing a $0.3$ difference from LSTM for seen and unseen patients, respectively. ``GluADFL(Cluster)'' is positioned intermediate between ``GluADFL(Random)'' and ``GluADFL(Ring)'' in terms of performance.
\end{itemize}

\section{Discussions}
\subsection{Cross prediction of population models towards unseen patients using federated learning}
\label{sec:cross_predict}

Recent advancements in deep learning have 
embraced methodologies such as transfer learning \cite{DBLP:journals/jhir/ZhuLCHG20, article_transfer, DBLP:journals/npjdm/ZhuULHOG22} and meta-learning \cite{9813400, langarica2023meta} to enhance the prediction of BG levels. These approaches have shown promise by optimizing population models through the application of insights gained across patient groups.

However, a notable drawback of transfer learning and meta-learning lies in their implications for patient privacy, necessitating the collection and aggregation of patient data. Additionally, both transfer learning and meta-learning require data from unseen patients for individual adaptation, conflicting with the preference for adaptation-free population models.


In the literature, population-based LSTM models have been recognized for their ability to accurately model the dynamics of BG levels across diverse patient groups, thereby delivering commendable outcomes in BGLP tasks \cite{DBLP:conf/ecai/BevanC20}. These models achieve this by discerning and applying general patterns across datasets without the need for individualized transfer adaptations. This characteristic enables them to effectively predict BG levels in patients not previously seen during training \cite{DBLP:conf/embc/MohebbiJHCTJBM20}. In our experiments, we also utilized four distinct datasets of varying population sizes, aim to validate the cross-prediction capabilities of population LSTMs for unseen patients, as detailed in Table \ref{table:gen_mixing}. The results demonstrate that these models, when trained and tested across different datasets, exhibit comparable performance. This consistency in performance extends to experiments involving ``GluADFL (Random)'', reinforcing the notion that population LSTMs, when trained through FL, retain their predictive accuracy in BGLP for unseen patients, as shown in Table \ref{table:gen_gluadfl}.

\begin{table*}[tb]
    \centering
    \caption{Blood Glucose Prediction for Seen/Unseen Patients by Different Population Methods on Four Datasets}
    \begin{threeparttable}
\resizebox{1.0\textwidth}{!}{
\begin{tabular}{|c|c|ccccccccccc|}
\hline
\multirow{2}{*}{Training Data} & \multirow{2}{*}{Metrics} & \multicolumn{11}{c|}{Methods Tested by \textbf{Seen} Patients}                                                                                                                              \\ \cline{3-13} 
                               &                          & LR          & XGBoost     & LSTM        & N-BEATS     & \multicolumn{1}{c:}{NHiTS}       & MAML        & MetaSGD     & \multicolumn{1}{c:}{FedAvg}      & GluADFL(Ring) & GluADFL(Cluster) & GluADFL(Random) \\ \hline
\multirow{5}{*}{OhioT1DM}      & RMSE                     & 21.48(5.38) & 20.64(3.72) & \textbf{19.65(3.59)} & {\color{red}  19.94(3.82)} & \multicolumn{1}{c:}{{\color{red}  19.89(3.75)}} & {\color{black}20.68(3.81)} & {\color{black}20.44(3.87)} & \multicolumn{1}{c:}{{\color{red}  19.65(3.60)}} & {\color{red}  19.74(3.63)}   & {\color{red}  19.65(3.58)}      & {\color{red}  19.66(3.59)}     \\
                               & MARD                     & 10.52(2.67) & 10.18(2.32) & \textbf{9.53(2.19)}  & {\color{red}  9.64(2.31)}  & \multicolumn{1}{c:}{{\color{red}  9.66(2.28)}}  & {\color{blue} 9.97(2.21)}  & {\color{black}10.23(2.42)} & \multicolumn{1}{c:}{{\color{red}  9.43(2.22)}}  & {\color{red}  9.52(2.24)}    & {\color{red}  9.47(2.24)}       & {\color{red}  9.43(2.23)}      \\
                               & MAE                      & 15.28(3.08) & 14.76(2.60) & \textbf{13.89(2.45)} & {\color{red}  14.08(2.58)} & \multicolumn{1}{c:}{{\color{red}  14.06(2.54)}} & {\color{black}14.77(2.61)} & {\color{black}14.74(2.65)} & \multicolumn{1}{c:}{{\color{red}  13.78(2.44)}} & {\color{red}  13.87(2.46)}   & {\color{red}  13.81(2.44)}      & {\color{red}  13.80(2.44)}     \\
                               & gRMSE                    & 26.94(7.44) & 26.07(5.16) & \textbf{24.65(5.02)} & {\color{blue} 25.00(5.34)} & \multicolumn{1}{c:}{{\color{red}  24.83(5.23)}} & {\color{black}25.81(5.37)} & {\color{blue} 25.10(5.32)} & \multicolumn{1}{c:}{{\color{red}  24.45(5.02)}} & {\color{red}  24.57(5.07)}   & {\color{red}  24.46(4.99)}      & {\color{red}  24.61(5.03)}     \\
                               & Time Lag                 & 7.17(4.90)  & 6.63(4.86)  & \textbf{5.11(4.29)}  & {\color{black}5.67(4.37)}  & \multicolumn{1}{c:}{{\color{black}5.76(4.42)}}  & {\color{black}5.88(4.39)}  & {\color{black}5.75(4.33)}  & \multicolumn{1}{c:}{{\color{red}  5.34(4.29)}}  & {\color{red}  5.40(4.29)}    & {\color{red}  5.41(4.30)}       & {\color{red}  5.22(4.28)}      \\ \hline
\multirow{5}{*}{ABC4D}         & RMSE                     & 27.14(4.40) & 23.56(3.60) & \textbf{22.42(3.33)} & {\color{blue} 22.92(3.45)} & \multicolumn{1}{c:}{{\color{red}  22.67(3.41)}} & {\color{black}22.99(3.43)} & {\color{black}23.23(3.49)} & \multicolumn{1}{c:}{{\color{red}  22.27(3.33)}} & {\color{blue} 22.73(3.43)}   & {\color{red}  22.42(3.36)}      & {\color{red}  22.28(3.33)}     \\
                               & MARD                     & 14.70(2.60) & 13.16(2.31) & \textbf{12.51(2.12)} & {\color{red}  12.64(2.17)} & \multicolumn{1}{c:}{{\color{red}  12.30(2.08)}} & {\color{red}  12.75(2.15)} & {\color{red}  12.54(2.06)} & \multicolumn{1}{c:}{{\color{red}  12.26(2.07)}} & {\color{red}  12.57(2.14)}   & {\color{red}  12.43(2.12)}      & {\color{red}  12.16(2.04)}     \\
                               & MAE                      & 18.68(3.06) & 16.45(2.56) & \textbf{15.69(2.33)} & {\color{red}  15.85(2.39)} & \multicolumn{1}{c:}{{\color{red}  15.64(2.34)}} & {\color{black}16.23(2.45)} & {\color{black}16.34(2.51)} & \multicolumn{1}{c:}{{\color{red}  15.44(2.30)}} & {\color{red}  15.77(2.37)}   & {\color{red}  15.57(2.33)}      & {\color{red}  15.41(2.30)}     \\
                               & gRMSE                    & 35.54(6.15) & 30.75(5.00) & \textbf{28.71(4.58)} & {\color{black}29.74(4.82)} & \multicolumn{1}{c:}{{\color{blue} 29.08(4.70)}} & {\color{black}29.50(4.74)} & {\color{black}30.06(4.92)} & \multicolumn{1}{c:}{{\color{red}  28.57(4.58)}} & {\color{blue} 29.03(4.69)}   & {\color{red}  28.77(4.62)}      & {\color{red}  28.60(4.59)}     \\
                               & Time Lag                 & 9.59(5.61)  & 7.93(4.75)  & \textbf{6.77(4.14)}  & {\color{red}  6.76(4.25)}  & \multicolumn{1}{c:}{{\color{red}  6.85(4.25)}}  & {\color{red}  6.60(4.20)}  & {\color{blue} 6.29(4.21)}  & \multicolumn{1}{c:}{{\color{red}  6.74(4.20)}}  & {\color{black}7.32(4.42)}    & {\color{red}  6.87(4.27)}       & {\color{red}  6.69(4.16)}      \\ \hline
\multirow{5}{*}{CTR3}          & RMSE                     & 22.95(4.67) & 20.83(3.59) & \textbf{20.02(3.39)} & {\color{red}  20.09(3.49)} & \multicolumn{1}{c:}{{\color{red}  20.13(3.44)}} & {\color{black}20.54(3.52)} & {\color{black}20.55(3.54)} & \multicolumn{1}{c:}{{\color{red}  20.00(3.40)}} & {\color{red}  20.30(3.49)}   & {\color{red}  20.07(3.42)}      & {\color{red}  20.00(3.40)}     \\
                               & MARD                     & 11.67(2.58) & {\color{blue}10.70(2.35)} & \textbf{10.24(2.17)} & {\color{red}  10.25(2.26)} & \multicolumn{1}{c:}{{\color{red}  10.29(2.21)}} & {\color{blue} 10.71(2.37)} & {\color{black}10.79(2.43)} & \multicolumn{1}{c:}{{\color{red}  10.08(2.12)}} & {\color{red}  10.28(2.17)}   & {\color{red}  10.16(2.17)}      & {\color{red}  10.09(2.12)}     \\
                               & MAE                      & 16.08(2.96) & 14.77(2.52) & \textbf{14.17(2.37)} & {\color{red}  14.17(2.39)} & \multicolumn{1}{c:}{{\color{red}  14.22(2.37)}} & {\color{blue} 14.65(2.46)} & {\color{black}14.69(2.47)} & \multicolumn{1}{c:}{{\color{red}  14.04(2.33)}} & {\color{red}  14.27(2.37)}   & {\color{red}  14.11(2.35)}      & {\color{red}  14.03(2.32)}     \\
                               & gRMSE                    & 28.53(6.54) & 25.81(5.01) & \textbf{24.70(4.75)} & {\color{red}  24.72(4.90)} & \multicolumn{1}{c:}{{\color{red}  24.60(4.79)}} & {\color{red}  25.00(4.89)} & {\color{red}  24.90(4.89)} & \multicolumn{1}{c:}{{\color{red}  24.47(4.75)}} & {\color{red}  24.70(4.85)}   & {\color{red}  24.56(4.79)}      & {\color{red}  24.42(4.74)}     \\
                               & Time Lag                 & 13.13(4.99) & 11.69(4.59) & \textbf{9.21(3.89)}  & {\color{red}  9.43(3.97)}  & \multicolumn{1}{c:}{{\color{blue} 9.60(4.02)}}  & {\color{black}10.19(4.18)} & {\color{black}9.96(4.11)}  & \multicolumn{1}{c:}{{\color{black} 9.72(4.02)}}  & {\color{black}10.48(4.23)}   & {\color{black}9.86(4.07)}       & {\color{black}9.76(4.04)}      \\ \hline
\multirow{5}{*}{REPLACE-BG}    & RMSE                     & 23.31(4.73) & 21.21(3.59) & \textbf{20.31(3.39)} & {\color{red}  20.43(3.51)} & \multicolumn{1}{c:}{{\color{red}  20.42(3.48)}} & {\color{black}21.82(3.68)} & {\color{black}21.44(3.61)} & \multicolumn{1}{c:}{{\color{red}  20.33(3.45)}} & {\color{black}20.87(3.64)}   & {\color{blue} 20.70(3.58)}      & {\color{red}  20.34(3.46)}     \\
                               & MARD                     & 11.85(2.54) & 11.18(2.27) & \textbf{10.50(2.09)} & {\color{red}  10.50(2.16)} & \multicolumn{1}{c:}{{\color{red}  10.43(2.09)}} & {\color{black}12.58(2.77)} & {\color{black}11.27(2.26)} & \multicolumn{1}{c:}{{\color{red}  10.43(2.11)}} & {\color{red}  10.75(2.23)}   & {\color{red}  10.69(2.21)}      & {\color{red}  10.45(2.12)}     \\
                               & MAE                      & 16.34(2.95) & 15.24(2.51) & \textbf{14.49(2.32)} & {\color{red}  14.50(2.37)} & \multicolumn{1}{c:}{{\color{red}  14.49(2.36)}} & {\color{black}16.32(2.73)} & {\color{black}15.60(2.58)} & \multicolumn{1}{c:}{{\color{red}  14.42(2.34)}} & {\color{blue} 14.80(2.46)}   & {\color{red}  14.70(2.43)}      & {\color{red}  14.44(2.34)}     \\
                               & gRMSE                    & 29.39(6.61) & 26.75(4.96) & \textbf{25.24(4.68)} & {\color{red}  25.46(4.88)} & \multicolumn{1}{c:}{{\color{red}  25.44(4.82)}} & {\color{black}27.99(5.21)} & {\color{black}27.40(5.17)} & \multicolumn{1}{c:}{{\color{red}  25.36(4.80)}} & {\color{black}25.98(5.05)}   & {\color{black} 25.85(4.98)}      & {\color{red}  25.38(4.81)}     \\
                               & Time Lag                 & 12.19(5.02) & 11.08(4.69) & \textbf{9.50(4.21)}  & {\color{red}  9.63(4.24) } & \multicolumn{1}{c:}{{\color{blue} 9.87(4.28)}}  & {\color{red}  8.40(4.05)}  & {\color{red}  8.88(4.12)}  & \multicolumn{1}{c:}{{\color{red}  9.45(4.22)}}  & {\color{black}10.49(4.48)}   & {\color{black}10.14(4.42)}      & {\color{red}  9.46(4.22)}      \\ 
\hline
\hline
\multirow{2}{*}{Training   Data} & \multirow{2}{*}{Metrics} & \multicolumn{11}{c|}{Methods Tested by ALL Other  \textbf{Unseen} Patients}                                                                                                                          \\ \cline{3-13} 
                                 &                          & LR          & XGBoost     & LSTM        & N-BEATS     & \multicolumn{1}{c:}{NHiTS}       & MAML        & MetaSGD     & \multicolumn{1}{c:}{FedAvg}      & GluADFL(Ring) & GluADFL(Cluster) & GluADFL(Random) \\ \hline
\multirow{5}{*}{OhioT1DM}            & RMSE                 & 24.50(5.38) & 21.72(3.72) & \textbf{20.99(3.59)} & {\color{black}21.52(3.82)} & \multicolumn{1}{c:}{{\color{black}21.41(3.75)}} & {\color{black}22.23(3.81)} & {\color{black}22.17(3.87)} & \multicolumn{1}{c:}{{\color{red}  20.97(3.60)}} & {\color{red}  21.08(3.63)}   & {\color{red}  20.94(3.58)}      & {\color{red}  20.96(3.59)}     \\
                                 & MARD                     & 12.16(2.67) & 11.51(2.32) & \textbf{10.96(2.19)} & {\color{red}  11.15(2.31)} & \multicolumn{1}{c:}{{\color{red}  11.14(2.28)}} & {\color{blue} 11.38(2.21)} & {\color{black}11.84(2.42)} & \multicolumn{1}{c:}{{\color{red}  10.89(2.22)}} & {\color{red}  10.95(2.24)}   & {\color{red}  10.91(2.24)}      & {\color{red}  10.88(2.23)}     \\
                                 & MAE                      & 16.60(3.08) & 15.57(2.60) & \textbf{14.94(2.45)} & {\color{red}  15.23(2.58)} & \multicolumn{1}{c:}{{\color{red}  15.17(2.54)}} & {\color{black}15.86(2.61)} & {\color{black}15.98(2.65)} & \multicolumn{1}{c:}{{\color{red}  14.85(2.44)}} & {\color{red}  14.92(2.46)}   & {\color{red}  14.85(2.44)}      & {\color{red}  14.85(2.44)}     \\
                                 & gRMSE                    & 30.99(7.44) & 27.52(5.16) & \textbf{26.37(5.02)} & {\color{black}27.04(5.34)} & \multicolumn{1}{c:}{{\color{blue} 26.82(5.23)}} & {\color{black}27.75(5.37)} & {\color{black}27.36(5.32)} & \multicolumn{1}{c:}{{\color{red}  26.19(5.02)}} & {\color{red}  26.35(5.07)}   & {\color{red}  26.15(4.99)}      & {\color{red}  26.30(5.03)}     \\
                                 & Time Lag                 & 11.60(4.90) & 11.25(4.86) & \textbf{9.28(4.29)}  & {\color{blue} 9.67(4.37)}  & \multicolumn{1}{c:}{{\color{black}9.93(4.42)}}  & {\color{blue} 9.72(4.39)}  & {\color{blue} 9.62(4.33)}  & \multicolumn{1}{c:}{{\color{red}  9.36(4.29)}}  & {\color{red}  9.33(4.29)}    & {\color{red}  9.37(4.30)}       & {\color{red}  9.25(4.28)}      \\ \hline
\multirow{5}{*}{ABC4D}           & RMSE                     & 23.69(4.40) & 21.35(3.60) & \textbf{20.39(3.33)} & {\color{red}  20.51(3.45)} & \multicolumn{1}{c:}{{\color{red}  20.43(3.41)}} & {\color{black}20.95(3.43)} & {\color{black}21.24(3.49)} & \multicolumn{1}{c:}{{\color{red}  20.17(3.33)}} & {\color{red}  20.46(3.43)}   & {\color{red}  20.26(3.36)}      & {\color{red}  20.18(3.33)}     \\
                                 & MARD                     & 12.53(2.60) & 11.38(2.31) & \textbf{10.75(2.12)} & {\color{red}  10.67(2.17)} & \multicolumn{1}{c:}{{\color{red}  10.51(2.08)}} & {\color{blue} 11.09(2.15)} & {\color{red}  10.99(2.06)} & \multicolumn{1}{c:}{{\color{red}  10.43(2.07)}} & {\color{red}  10.63(2.14)}   & {\color{red}  10.52(2.12)}      & {\color{red}  10.36(2.04)}     \\
                                 & MAE                      & 17.21(3.06) & 15.46(2.56) & \textbf{14.69(2.33)} & {\color{red}  14.66(2.39)} & \multicolumn{1}{c:}{{\color{red}  14.56(2.34)}} & {\color{black}15.28(2.45)} & {\color{black}15.45(2.51)} & \multicolumn{1}{c:}{{\color{red}  14.40(2.30)}} & {\color{red}  14.64(2.37)}   & {\color{red}  14.48(2.33)}      & {\color{red}  14.37(2.30)}     \\
                                 & gRMSE                    & 30.12(6.15) & 27.14(5.00) & \textbf{25.43(4.58)} & {\color{blue} 25.77(4.82)} & \multicolumn{1}{c:}{{\color{red}  25.39(4.70)}} & {\color{black}26.19(4.74)} & {\color{black}26.85(4.92)} & \multicolumn{1}{c:}{{\color{red}  25.11(4.58)}} & {\color{red}  25.31(4.69)}   & {\color{red}  25.21(4.62)}      & {\color{red}  25.18(4.59)}    \\
                                 & Time Lag                 & 14.01(5.61) & 11.15(4.75) & \textbf{9.39(4.14)}  & {\color{red}  9.62(4.25)}  & \multicolumn{1}{c:}{{\color{blue} 9.70(4.25)}}  & {\color{red}  9.18(4.20)}  & {\color{red}  9.10(4.21)}  & \multicolumn{1}{c:}{{\color{red}  9.47(4.20)}}  & {\color{black}10.28(4.42)}   & {\color{red}  9.69(4.27)}       & {\color{red}  9.42(4.16)}      \\ \hline
\multirow{5}{*}{CTR3}            & RMSE                     & 23.72(4.67) & 21.39(3.59) & \textbf{20.74(3.39)} & {\color{red}  20.78(3.49)} & \multicolumn{1}{c:}{{\color{red}  20.71(3.44)}} & {\color{blue} 21.16(3.52)} & {\color{blue} 21.20(3.54)} & \multicolumn{1}{c:}{{\color{red}  20.52(3.40)}} & {\color{red}  20.73(3.49)}   & {\color{red}  20.61(3.42)}      & {\color{red}  20.49(3.40)}     \\
                                 & MARD                     & 12.23(2.58) & {\color{blue} 11.47(2.35)} & \textbf{10.99(2.17)} & {\color{red}  10.97(2.26)} & \multicolumn{1}{c:}{{\color{red}  10.94(2.21)}} & {\color{black}11.51(2.37)} & {\color{black}11.65(2.43)} & \multicolumn{1}{c:}{{\color{red}  10.65(2.12)}} & {\color{red}  10.79(2.17)}   & {\color{red}  10.79(2.17)}      & {\color{red}  10.64(2.12)}     \\
                                 & MAE                      & 16.67(2.96) & {\color{blue} 15.40(2.52)} & \textbf{14.91(2.37)} & {\color{red}  14.84(2.39)} & \multicolumn{1}{c:}{{\color{red}  14.81(2.37)}} & {\color{blue} 15.29(2.46)} & {\color{blue} 15.37(2.47)} & \multicolumn{1}{c:}{{\color{red}  14.56(2.33)}} & {\color{red}  14.71(2.37)}   & {\color{red}  14.67(2.35)}      & {\color{red}  14.54(2.32)}     \\
                                 & gRMSE                    & 30.17(6.54) & 27.26(5.01) & \textbf{26.30(4.75)} & {\color{red}  26.27(4.90)} & \multicolumn{1}{c:}{{\color{red}  26.00(4.79)}} & {\color{red}  26.47(4.89)} & {\color{red} 26.41(4.89)}  & \multicolumn{1}{c:}{{\color{red}  25.80(4.75)}} & {\color{red}  25.92(4.85)}   & {\color{red}  25.92(4.79)}      & {\color{red}  25.72(4.74)}     \\
                                 & Time Lag                 & 11.48(4.99) & 10.10(4.59) & \textbf{8.15(3.89)}  & {\color{red}  8.23(3.97)}  & \multicolumn{1}{c:}{{\color{red}  8.41(4.02)}}  & {\color{black}8.90(4.18)}  & {\color{black}8.69(4.11)}  & \multicolumn{1}{c:}{{\color{blue} 8.53(4.02)}}  & {\color{black}9.13(4.23)}    & {\color{blue} 8.64(4.07)}       & {\color{blue} 8.59(4.04)}      \\ \hline
\multirow{5}{*}{REPLACE-BG}      & RMSE                     & 24.69(4.73) & 21.93(3.59) & \textbf{21.04(3.39)} & {\color{red}  21.24(3.51)} & \multicolumn{1}{c:}{{\color{red}  21.20(3.48)}} & {\color{black}22.17(3.68)} & {\color{black}21.97(3.61)} & \multicolumn{1}{c:}{{\color{red}  21.13(3.45)}} & {\color{black}21.78(3.64)}   & {\color{black}21.56(3.58)}      & {\color{red}  21.14(3.46)}     \\
                                 & MARD                     & 12.45(2.54) & 11.52(2.27) & \textbf{10.92(2.09)} & {\color{red}  10.96(2.16)} & \multicolumn{1}{c:}{{\color{red}  10.85(2.09)}} & {\color{black}12.70(2.77)} & {\color{black}11.58(2.26)} & \multicolumn{1}{c:}{{\color{red}  10.89(2.11)}} & {\color{blue} 11.27(2.23)}   & {\color{red}  11.18(2.21)}      & {\color{red}  10.91(2.12)}     \\
                                 & MAE                      & 16.75(2.95) & 15.40(2.51) & \textbf{14.70(2.32)} & {\color{red}  14.75(2.37)} & \multicolumn{1}{c:}{{\color{red}  14.72(2.36)}} & {\color{black}16.17(2.73)} & {\color{black}15.61(2.58)} & \multicolumn{1}{c:}{{\color{red}  14.68(2.34)}} & {\color{blue} 15.14(2.46)}   & {\color{red}  15.00(2.43)}      & {\color{red}  14.69(2.34)}     \\
                                 & gRMSE                    & 31.27(6.61) & 27.69(4.96) & \textbf{26.27(4.68)} & {\color{blue} 26.62(4.88)} & \multicolumn{1}{c:}{{\color{red}  26.53(4.82)}} & {\color{black}28.39(5.21)} & {\color{black}28.09(5.17)} & \multicolumn{1}{c:}{{\color{red}  26.48(4.80)}} & {\color{black}27.23(5.05)}   & {\color{black}27.03(4.98)}      & {\color{red}  26.50(4.81)}     \\
                                 & Time Lag                 & 10.52(5.02) & 9.56(4.69)  & \textbf{8.13(4.21)}  & {\color{red}  8.18(4.24)}  & \multicolumn{1}{c:}{{\color{blue} 8.44(4.28)}}  & {\color{red}  7.18(4.05)}  & {\color{red}  7.60(4.12)}  & \multicolumn{1}{c:}{{\color{red}  8.09(4.22)}}  & {\color{black}9.00(4.48)}    & {\color{black}8.68(4.42)}       & {\color{red}  8.12(4.22)}      \\ \hline
\end{tabular}
}
    \begin{tablenotes}
   
    \item {\fontsize{5.5}{1.5}\selectfont \textbf{Bold}: results for LSTM trained by supervised learning;  \par}
    \item {\fontsize{5.5}{1.5}\selectfont {\color{red} Red}: differences between LSTM trained by supervised learning and other methods are below $0.30$ for each row;  \par}
    \item {\fontsize{5.5}{1.5}\selectfont {\color{blue} Blue}: differences between LSTM trained by supervised learning and other methods are between $(0.3, 0.5]$ for each row.  \par}
    
    \end{tablenotes}
  \end{threeparttable}
    \label{table:compare_seen_unseen}
\end{table*}

Additionally, seen patients whose data are included in the training set can further refine and personalize the population model post FL training, as illustrated in Figure \ref{fig:fine_tune}. Such customization is achieved by adjusting the model parameters to better fit the individual's unique data, thereby enhancing the model's accuracy and effectiveness for their specific BG level predictions.
However, the distinction between ``Population Model'' and ``Personalized from Population'' is slight, highlighting population features are more important.
Therefore, we opt not to focus on personalized FL in current work.



\subsection{The impact of different communication typologies of GLuADFL on BGLP}

We explore three sparse communication topologies for decentralized FL, specifically ring, cluster, and random, as suggested in \cite{DBLP:conf/icml/Dai0H0T22, DBLP:conf/iclr/BornsteinRWBH23}. In Figure \ref{fig:rmse_converg}, with a communication batch size of $B=7$, we observe some differences in the training efficacy of decentralized FL based on the communication graph topology. The random topology demonstrates the best convergence, i.e., converging at the lowest RMSE, followed by the cluster topology, with the ring topology showing relatively the worst convergence.

This pattern can be attributed to the number of connections in each topology; more connections typically result in better convergence, i.e., converging at lower RMSE. The cluster topology, with its moderate number of connections, converges better than the ring topology but worse than the random topology. The random topology's time-varying connections facilitate wider broadcasting of model parameters, leading to its best convergence.

Moreover, as the population size increases (from OhioT1DM to REPLACE-BG), the differences in convergence performance among these topologies become more notable. Specifically, topologies with fewer connections, like the ring topology, tend to transmit model parameters more slowly and limited during communications.

Additionally, we implement a wait-free mechanism, as discussed in \cite{DBLP:conf/iclr/BornsteinRWBH23}, where only active nodes participate in communication and local model updates. Figure \ref{fig:rmse_inactive} reveals the impact of inactive node ratios on the prediction performance across different communication topologies. In line with our earlier observations, the random topology maintains stability even as the inactive node ratio increases. Notably, larger datasets amplify the advantages of the random topology, showing more significant performance gaps compared to other topologies. Figure \ref{fig:rmse_inactive} also indicates a sharp decline in the effectiveness of GluADFL when the inactive node ratio exceeds 70\%.

These observations reveal that our GluADFL framework adapts well to varying network topologies, with the random topology showing particular resilience in unstable or dynamic network conditions. Its robust performance, even with a high inactive node ratio, demonstrates its potential for real-world applications where network scenarios are variable for different social preferences. The observed scalability with larger datasets reinforces GluADFL’s suitability for extensive decentralized learning environments. 

\subsection{Comparison of blood glucose prediction performance}

In comparing GluADFL with baselines on four datasets for both seen and unseen patients (see Section \ref{sec:baselines}, Tables \ref{table:compare_seen_unseen}), we observed the following:

1) LSTM models outperform traditional algorithms like LR and XGBoost in BGLP. Despite the effectiveness of LR and XGBoost in BGLP as documented in \cite{articlelr, articlexgb}, they fall short in comparison to LSTM in all metrics for both seen and unseen patients. Advanced deep learning models like N-BEATS and NHiTS, though excellent in healthcare applications \cite{DBLP:conf/ecai/Rubin-FalconeFW20, potosnak2023forecasting}, do not surpass LSTM when using only BG data as input, similar to findings in \cite{puszkarski2022comparison}. This could be because N-BEATS and NHiTS are more suited for complex time series forecasting than single point predictions.

2) Population FL-based LSTMs have inherent generalization and cross-predictive capabilities for unseen patients, as seen in section \ref{sec:cross_predict}. We test meta-learning approaches like MAML and MetaSGD, known for quick adaptation to new tasks, in our baseline models without fine-tuning for unseen patients. However, these meta-learning models cannot outperform LSTM, FedAvg, or GluADFL for unseen patients. Consequently, we decide against incorporating a meta-learning module in GluADFL and did not pursue meta-learning-based FL \cite{Chen2018FederatedMW} or domain generation-based FL \cite{DBLP:conf/cvpr/ZhangXYZ0W23} further.

3) GluADFL shows comparable performance to FedAvg, indicating that the FL and decentralized structure do not compromise LSTM's modeling and cross-predictive abilities. GluADFL also addresses the ``cold start'' problem in BGLP, enabling cross-prediction for unseen patients while protecting privacy. Therefore, we choose not to add complex modules to GluADFL, such as masked model-parameter averaging \cite{DBLP:conf/icml/Dai0H0T22} or client-communication weight selection \cite{DBLP:conf/iclr/BornsteinRWBH23}.

\section{Limitations and Future work}
This study, while contributing valuable insights into BGLP using FL and population LSTM models, acknowledges certain limitations that present areas for future exploration.

Our current investigation is confined to single-horizon BG level predictions, and does not extend to multi-horizon forecasting. Multi-horizon prediction involves estimating BG levels at multiple future time points, offering a more comprehensive view of glucose fluctuations over time. This limitation highlights a significant area for further research, where expanding the scope to include multi-horizon predictions could enhance the utility of our models for more dynamic and anticipatory diabetes management.

The exploration of our proposed method is focused primarily on BGLP and has yet to encompass other critical areas of diabetes management, such as insulin dose recommendation, dietary advice, or physical activity impact analysis. The complexity and multifaceted nature of diabetes care suggest a wealth of opportunities for applying our methodology to other aspects of diabetes management, potentially offering a more holistic approach to managing the condition.

The focus of this work is on designing an FL training framework. Therefore, we adopted LSTM as a basic deep learning-based approach in BGLP and will more advanced models like those based on transformers \cite{DBLP:conf/nips/VaswaniSPUJGKP17, DBLP:conf/iscas/ZhuCKZLG23, 10379014, DBLP:conf/icassp/SergazinovAG23, DBLP:journals/bdcc/AcunaAP23}, in the future.


\section{Conclusions}
To address the ``cold start'' problem in BGLP while ensuring privacy, we propose GluADFL, an asynchronous decentralized FL framework. 
We employed LSTMs as a practical and efficient solution for individual patient BG level predictions. 
Using four T1D datasets, we demonstrated GluADFL's ability to maintain LSTM's cross-predictive capacity, showing scalability with larger populations and robustness across various communication topologies and asynchronous settings. 
Hence, we successfully established a robust and privacy-preserving approach to BGLP, offering both high accuracy and adherence to privacy considerations.


\section{Acknowledgement} 
We thank the support from UKRI Center for Doctoral Training in AI-enabled healthcare systems [EP/S021612/1] and University College London Overseas Research Scholarships.
{\color{black}
We also appreciate the support from Rosetrees Trust (Grant number: UCL-IHE-2020\textbackslash102) and Great Ormond Street Hospital (Charity ref.X12018).
The study sponsors had no study involvement.
}



\bibliographystyle{elsarticle-num-names}
\bibliography{ref}


\end{document}